\documentclass[runningheads]{llncs}

\usepackage{eccv}

\usepackage{eccvabbrv}

\usepackage{graphicx}
\usepackage{booktabs}

\usepackage{tcolorbox}
\usepackage{makecell}
\usepackage[linesnumbered,ruled]{algorithm2e}
\usepackage{colortbl}
\usepackage{float}
\usepackage{stfloats}
\usepackage{enumitem}
\usepackage{booktabs}
\usepackage{multirow}
\usepackage{threeparttable}
\usepackage{amssymb}
\usepackage{adjustbox}
\usepackage{lineno}
\usepackage[x11names]{xcolor}
\usepackage{caption}
\usepackage{amsmath}
\usepackage{marvosym}

\usepackage[accsupp]{axessibility}  

\usepackage[breaklinks,colorlinks,citecolor=eccvblue]{hyperref}

\usepackage{orcidlink}

\begin{document}

\title{Keeping the Evidence Chain: Semantic Evidence Allocation for Training-Free Token Pruning in Video Temporal Grounding} 

\titlerunning{Keeping the Evidence Chain}


\author{
Jiaqi Li\inst{1}\orcidlink{0000-0001-8509-5528} \and
Shuntian Zheng\inst{1}\orcidlink{0000-0002-5717-5851} \and
Yixian Shen\inst{2}\orcidlink{0000-0001-8447-872X} \and
Jia-Hong Huang\inst{2,3}\orcidlink{0000-0001-7943-2591} \and
Xiaoman Lu\inst{1}\orcidlink{0009-0004-9839-1405} \and
Minzhe Ni\inst{1}\orcidlink{0009-0008-8553-6221} \and
Yu Guan\inst{1}\orcidlink{0000-0002-1283-3806}\textsuperscript{(\Letter)}
}

\authorrunning{J. Li et al.}

\institute{University of Warwick, Coventry CV4 7AL, United Kingdom \\
\email{
\{jiaqi.li.16, yu.guan\}@warwick.ac.uk
}
\and
University of Amsterdam, Amsterdam 1012 WX, Netherlands
\and
Amazon AGI, The United States of America
\\
}

\maketitle

\definecolor{lightgray}{gray}{0.93}

\begin{abstract}
Video Temporal Grounding (VTG) localizes the temporal boundaries of query-relevant moments in long, untrimmed videos, making video-language-model prohibitively expensive. 
While recent training-free token pruning has shown success in video question answering, naively applying these objectives to VTG causes drastic degradation, as VTG crucially depends on boundary-sensitive evidence and cross-frame reasoning chains. 
We therefore identify two VTG-specific pruning principles: evidence retention, which keeps query-critical patches especially around event boundaries, and connectivity strength, which preserves cross-frame connectivity for long-range evidence aggregation. 
Building on these insights, we propose SemVID, a training-free pruning framework that constructs a compact yet coherent token subset with complementary semantic roles. 
SemVID first allocates per-frame budgets by balancing query relevance and inter-frame variation to avoid over-pruned segments, and then selects three types of tokens: object tokens for diverse query-critical evidence, motion tokens to capture meaningful transitions and serve as cross-frame relays, and context tokens for scene continuity. 
Extensive experiments show that SemVID achieves a strong accuracy-efficiency trade-off, retaining up to 95.4\% mIoU with only 12.5\% visual tokens and delivering up to a 5.8× prefill speedup, consistently outperforming prior methods under the same budgets.
Our code is available \href{https://github.com/JiaqiLi404/SemVID}{here}.

\end{abstract}  
\section{Introduction}
\label{sec:intro}

Video Temporal Grounding (VTG) aims to localize the start and end timestamps of a moment in an untrimmed video that matches a language query \cite{chen2021end,qu2024chatvtg,zheng2024training,wang2023protege}. 
As a core capability for practical video interaction, VTG supports moment retrieval, highlight discovery, and query-driven video summarization, where users need to quickly jump to the exact moment of interest \cite{mun2020local,lin2023univtg}. 
Recently, VTG methods have begun to leverage Video-Language Models (VLMs), benefiting from their strong cross-modal understanding and reasoning ability, and have achieved promising performance on diverse VTG benchmarks \cite{lin2023univtg,chen2025localizing}.

Despite recent progress, deploying VLM-based VTG remains expensive.
A video is typically tokenized into thousands of patch tokens, and the attention cost scales quadratically with sequence length \cite{dosovitskiy2020image,vaswani2017attention,shinde2025survey}.
This challenge is amplified for long videos: precise boundary localization often requires dense sampling \cite{wang2025time}, yet increasing the sampling rate quickly makes prefill consumption prohibitive.

As described in \cite{kim2022learned,kim2024token}, many visual tokens are redundant and contribute little to performance and thus can be safely removed without compromising accuracy.
This naturally motivates training-free visual token pruning to reduce computation.
However, pruning strategies designed for VTG remain under-explored.
In practice, existing training-free pruning baselines are borrowed from Video Question Answering (VideoQA) and can be categorized by their objectives into Visual Redundancy (VR), Visual Saliency (VS), and Query Relevance (QR).

\begin{figure}[tb]
  \centering
  \includegraphics[width=\linewidth]{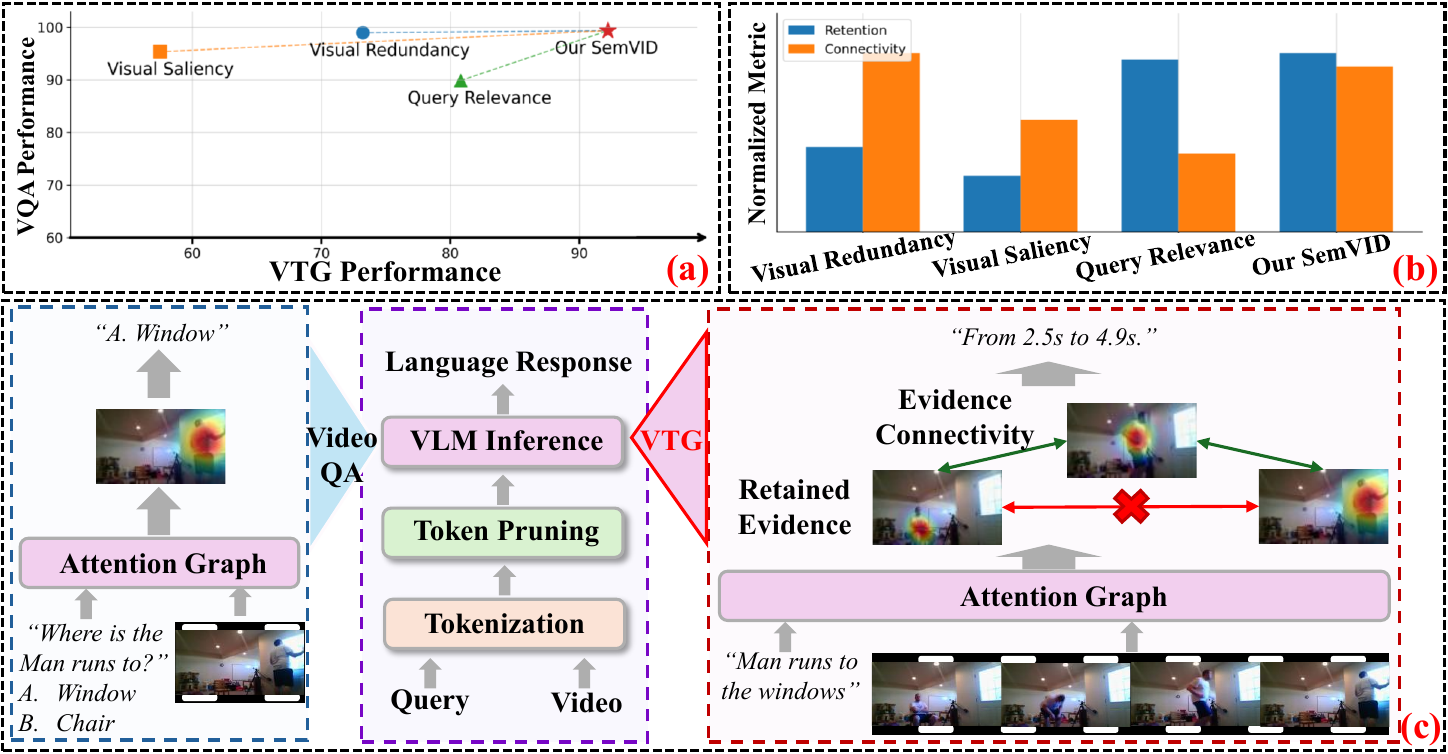}
  \caption{
Comparison between existing pruning objectives and SemVID for VTG.
\textbf{(a)} Performance comparison between VTG and VideoQA tasks.
\textbf{(b)} Diagnostics of pruning objectives on evidence retention and cross-frame connectivity.
\textbf{(c)} VTG requires long-range evidence aggregation rather than a single informative frame. 
SemVID preserves both query-critical evidence and transition relays to connect evidence across frames. 
}
\label{fig:motivation}
\end{figure}

A straightforward attempt is to directly apply these VideoQA pruning objectives to VTG.
While effective for perception-oriented tasks (e.g., object/attribute recognition) that can often be answered from a single informative frame \cite{wu2024longvideobench,fu2025video,lei2023revealing}, VTG fundamentally differs: it requires temporally coherent evidence to localize event boundaries and to reason about how events evolve over time \cite{chen2025localizing,wu2025survey}.
As a result, naively transferring VideoQA pruning tends to discard temporally critical cues and leads to severe performance drops, as shown in \cref{fig:motivation}(a).

This mismatch can be understood through the lens of VTG-specific requirements.
VR removes duplicate content by merging or discarding visually similar tokens. 
While effective for compression, its query-agnostic criterion can suppress small yet decisive evidence near event boundaries \cite{bolya2022tome}. 
VS prioritizes salient regions, but saliency often concentrates tokens on a few standout frames, leaving large portions of the timeline underrepresented \cite{shen2025fastvid}. 
This produces insufficient temporal coverage and weakens the evidence continuity, making it difficult to track evolving events. 
QR keeps tokens most similar to the query. 
However, it often repeatedly selects the same locally relevant regions, producing fragmented glimpses that miss the state transitions and disrupt cross-frame reasoning \cite{kumar2025lgttp,fastv}.

Motivated by these observations, we argue that effective pruning for VTG should satisfy two objectives: \textbf{Evidence Retention (ER)} and \textbf{Connectivity Strength (CS)}.
ER aims to preserve query-critical evidence patches, especially those around temporal boundaries.
CS requires that the retained tokens remain well connected across frames so evidence can be aggregated along the video timeline.
From an attention-graph perspective, query-conditioned signals are extracted through cross-attention and propagated across frames and layers via stacked self-attention as multi-hop message passing \cite{abnar2020quantifying}.
Pruning alters the graph topology by removing nodes and edges.
If boundary-critical evidence tokens or intermediate relay tokens are dropped, the evidence chain becomes fragmented, impeding long-range temporal aggregation and degrading grounding accuracy \cite{chefer2021transformer,zheng2025reasoning}.
The more details are demonstrated in \cref{fig:motivation}(c).


To address the aforementioned challenges, we propose \textbf{SemVID}, a training-free pruning framework tailored for VTG. 
SemVID explicitly optimizes ER and CS by constructing a compact set of tokens with complementary semantic roles. 
Concretely, SemVID first assigns each frame a token budget that balances query relevance and inter-frame variation, preventing empty or over-pruned segments in long videos. 
It then identifies three types of tokens:
(i) \textbf{object tokens} that preserve diverse query-aligned evidence for ER; 
(ii) \textbf{motion tokens} that capture meaningful temporal changes and act as cross-frame relays for CS; 
and (iii) a small number of \textbf{context tokens} as stable anchors to maintain scene continuity. 
As reflected in \cref{fig:motivation}(b), SemVID achieves strong performance on both ER and CS, and accordingly, these role-aware tokens form a compact yet coherent evidence chain that keeps evidence both present and traceable while substantially reducing visual tokens (\cref{fig:motivation}c).
Extensive experiments demonstrate that SemVID achieves a strong accuracy-efficiency trade-off on VTG, retaining up to 95.4\% mIoU with only 12.5\% tokens while delivering a 5.8$\times$ prefill speedup, outperforming prior training-free pruning objectives under the same token budget.

Our contributions are as follows:
\begin{itemize}[itemsep=0pt,topsep=0pt,parsep=0pt,leftmargin=10pt]
\item We identify that VTG-oriented pruning should follow objectives, Evidence Retention (ER) and Connectivity Strength (CS), which are crucial for preserving boundary-critical evidence and maintaining cross-frame reasoning chains.
\item We propose SemVID, a training-free pruning framework tailored for VTG, which explicitly optimizes ER and CS by constructing a role-aware token set that preserves query-critical evidence and maintains cross-frame connectivity.
\item We demonstrate a strong accuracy-efficiency trade-off on VTG benchmarks, consistently outperforming prior methods under the same token budgets.
\end{itemize}

\section{Related Work}
\label{sec:related}

\textbf{Training-Free Pruning for VLMs.}
Modern VLMs tokenize videos into dense patch tokens \cite{dosovitskiy2020image,jiang2026imagine}, resulting in long visual sequences and quadratic attention cost \cite{vaswani2017attention}.
This motivates training-free token pruning to accelerate \cite{yang2025visionzip}.
However, pruning for VTG is particularly challenging.
Unlike coarse VideoQA, VTG requires long-range evidence that is both retained and temporally traceable \cite{chen2025localizing,wu2025survey}.  
Aggressive pruning can appear safe on coarse QA yet fail on localizations \cite{endo2025feather}.


\textbf{Visual Redundancy.}
VR reduces duplicate content by merging or discarding redundant tokens.
PruneVid \cite{huang2025prunevid} clusters frames into scenes and compresses static tokens within each scene.
FastVID \cite{shen2025fastvid} further preserves spatiotemporal structure during compression.
ToMe \cite{bolya2022tome} merges similar tokens around fixed anchors.
TokenSculpt \cite{qatokensculpt} introduces structure-aware merging to better respect video geometry.
However, VR is typically query-agnostic and tends to remove boundary-sensitive transition cues.
We mitigate this via role-aware token allocation, preserving query-critical objects and transitions to safeguard evidence.

\textbf{Visual Saliency.}
VS ranks tokens by saliency or attention and keeps the top ones \cite{fastv,shen2025fastvid,zhang2025vispruner,liu2024revisiting}.
FastVID computes saliency by vision-encoder attention \cite{shen2025fastvid}.
However, attention-based scores can be unstable and biased by attention sinks \cite{xiao2023efficient,li2026towards}, 
leading to the selected tokens not corresponding to semantically informative regions \cite{huang2024ivtp,yang2025visionzip,dart}.
Importantly, VS often concentrates tokens on a few dominant frames and causes token-empty gaps that break evidence chains.
We address this with a lightweight budget-allocation prior that reserves per-frame tokens to maintain temporal coverage and continuity.

\textbf{Query Relevance.}
QR conditions retention on the query, typically by query-token similarity or cross-modal attention.
PruneVid combines redundancy reduction with question-guided selection \cite{huang2025prunevid}.
IVTP first estimates intra-vision importance and then filters tokens with instruction-related semantics \cite{huang2024ivtp}.
LGTTP increases token density for temporally relevant segments based on query cues \cite{kumar2025lgttp}.
Despite their appeal, attention-based relevance can be biased by attention sinks and may even underperform simple baselines \cite{zhang2025vispruner,wen2025token,dart}.
Furthermore, it tends to extract scattered local patches across frames, leading to fragmented evidence and weakened cross-frame connectivity for multi-hop reasoning \cite{abnar2020quantifying}.
To avoid sink-induced bias, we use simple query-token similarity, whose effectiveness has been widely validated in other fields \cite{radford2021learning, lewis2020retrieval,reimers2019sentence}.
Moreover, we promote diversity to avoid repeatedly selecting the same regions, reducing fragmented glimpses.

\section{SemVID}
\begin{figure}[tb]
  \centering
  \includegraphics[width=\linewidth]{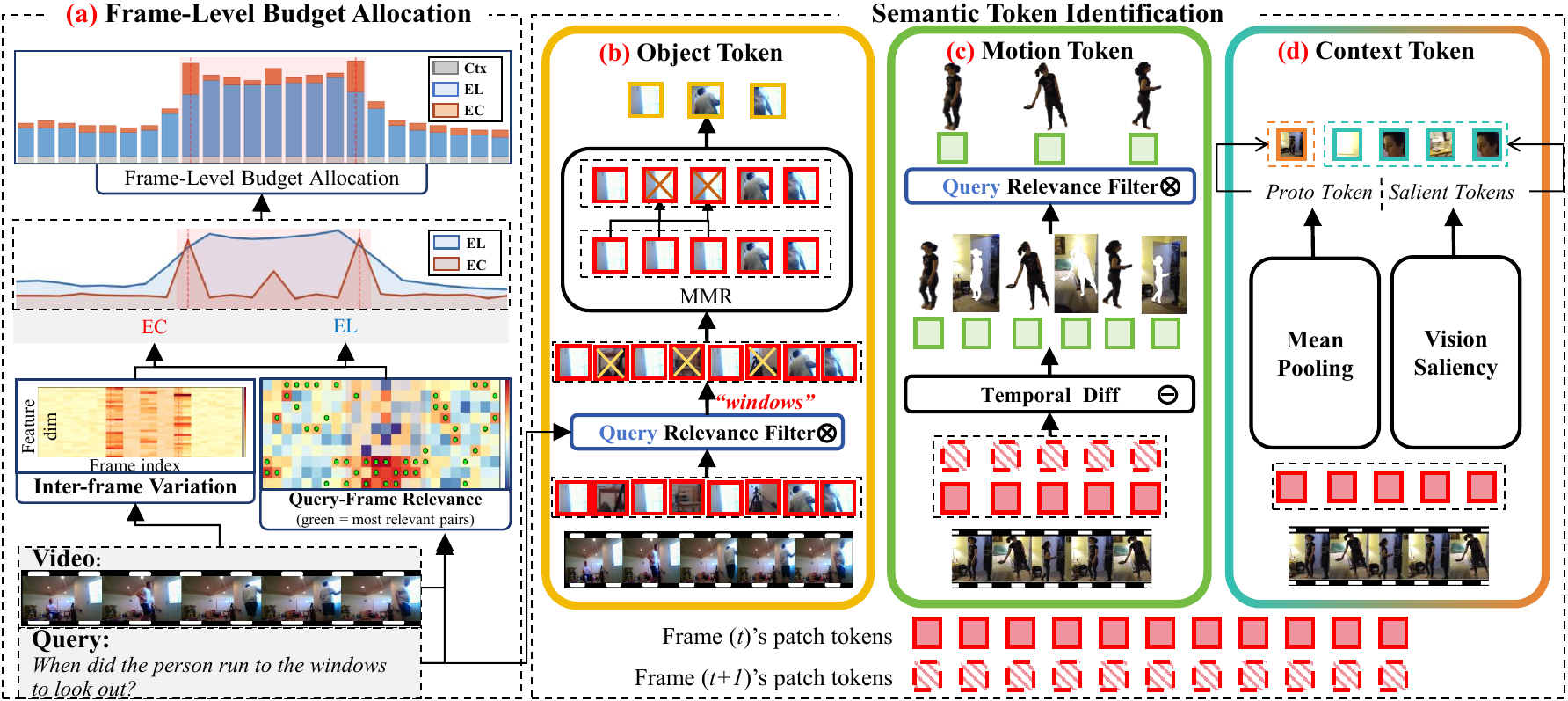}
  \caption{
Overview of SemVID semantic-oriented pruning.
(a) \textbf{Frame-level budget allocation}: assigns per-frame token budgets by jointly considering query-frame relevance and inter-frame variation. Given per-frame budgets, SemVID then outputs three roles of tokens.
(b) \textbf{Object token}: uses Maximal Marginal Relevance (MMR) to retain query-relevant yet diverse evidence.
(c) \textbf{Motion token}: retain query-aligned transitions as relay nodes to bridge long-range evidence and preserve connectivity.
(d) \textbf{Context token}: selects per-frame anchors by scene-level representativeness and saliency.
}
  \label{fig:overview}
\end{figure}

\subsection{Problem Definition}
We consider a Video-Language Model (VLM) that encodes a video into patch tokens and then performs question-answering with a language query.
Given a video with $T$ frames, each frame is tokenized into $P$ visual tokens, producing token embeddings
$V_{\text{patch}}\in\mathbb{R}^{T\times P\times D}$, where $D$ is the dimension of hidden states.
We also compute a frame-level global feature by applying mean pooling to the patch dimension, forming $V_{\text{glb}}\in\mathbb{R}^{T\times D}$.
The query is represented by token embeddings $Q\in\mathbb{R}^{N\times D}$, where $N$ is the query token length.
For VLM processing, the queries are combined with different instructions depending on the task.

\textbf{Video Temporal Grounding.}
Given a video and a query describing an event, VTG aims to localize the start and end timestamps of that event.

\textbf{SemVID Overview.} Given a retention ratio $r$, we select a subset of visual tokens $\mathcal{V}'$ such that $|\mathcal{V}'|= r\!\cdot\!(TP)$ without compromising performance. 
As shown in \cref{fig:overview}, SemVID first performs frame-level budget allocation to distribute the budget across $T$ frames, producing per-frame token quotas $\{k^{(t)}\}_{t=1}^{T}$. 
Given these budgets, we then conduct role-aware semantic token selection within each frame. 
Since object tokens localize query-critical evidence, we first allocate $\alpha\!\cdot\!k^{(t)}$ slots to object tokens, where $\alpha$ balances evidence preservation and connectivity.
We then use the remaining $(1-\alpha)k^{(t)}-k_{\text{ctx}}$ slots for motion tokens to bridge evidence and finally reserve $k_{\text{ctx}}$ context tokens as auxiliary anchors.

\subsection{Frame-Level Budget Allocation}
SemVID balances two objectives derived from the attention graph: 
(1) \emph{evidence localization}, which prioritizes frames where the query injects evidence; and
(2) \emph{evidence connectivity}, which emphasizes transition-rich frames that serve as temporal relays between evidence-bearing moments.
Relying only on query relevance tends to concentrate the budget on a few locally discriminative frames, producing isolated evidence snapshots.
By also allocating tokens to transition-rich frames, SemVID preserves the intermediate state changes that make these evidence snapshots temporally traceable.
This connectivity-aware allocation is particularly important for VTG, as precise boundary localization depends on linking evidence through intermediate state changes rather than fragment frames.
More discussion on these objectives is recorded in Appx. A.1.

\textbf{Evidence Localization.}
To quantify which frames are more likely to contain query-critical evidence, we use a lightweight query-relevance over frame features:
\begin{equation}
\label{eq:evidence_localization}
s_{\text{EL}}^{(t)}
=
\frac{1}{N}\cdot \hat{V}_{\text{glb}}^{(t)} \hat{Q}^\top,
\end{equation}
where $\hat{V}_{\text{glb}}^{(t)}\in\mathbb{R}^{D}$ and $\hat{Q}\in\mathbb{R}^{N\times D}$ denote normalized features of each frame and query tokens.
This score provides a lightweight estimate of where query evidence is injected. 
Since both frame and query features are normalized, high similarity indicates frames whose global semantics are more aligned with the query.

\textbf{Evidence Connectivity.}
We also allocate budget to transition-rich frames critical for connecting long-range evidence flow.
A cheap proxy is used to localize intermediate frames by measuring the temporal change of frames:
\begin{equation}
\label{eq:ivm}
 s_{\text{EC}}^{(t)}= \begin{cases}
 \|\hat V_{\text{glb}}^{(t)}-\hat V_{\text{glb}}^{(t-1)}\|_2, & t\ge 2,\\[6pt]
 s_{\text{EC}}^{(2)}, & t=1.
 \end{cases}
\end{equation}
Large $s_{\text{EC}}^{(t)}$ suggests a likely state transition, which serves as intermediate evidence needed for connectivity.
We allocate more tokens to such frames to preserve transition cues and keep the multi-hop evidence path continuous.

\textbf{Budget Allocation.}
Given a retention ratio $r$, we keep $|\mathcal{V}'|=r\cdot(TP)$ tokens in total.
We compute a mixed per-frame weight by
\begin{equation}
\label{eq:ba_w}
 w^{(t)}=\alpha\, s_{\text{EL}}^{(t)} + (1-\alpha)\, s_{\text{EC}}^{(t)}.
\end{equation}
We reuse $\alpha$ to balance query relevance and transitions.
To avoid token-empty frames, we use a per-frame floor $k_{\text{ctx}}$.
The final per-frame budget is
\begin{equation}
\label{eq:ba_k}
 k^{(t)} = (K-Tk_{\text{ctx}})\cdot \frac{w^{(t)}}{\sum_{i=1}^{T} w^{(i)}} + k_{\text{ctx}}.
\end{equation}
This allocation mainly follows the weight $w^{(t)}$, while $k_{\text{ctx}}$ serves as an auxiliary safeguard for, avoiding token-empty gaps that would break temporal continuity.

\subsection{Object Token}
To localize evidence, a natural strategy is to retain the tokens that the query attends to. 
However, attention-based query relevance methods that compute cross-attention over all patches require materializing large attention maps, which introduces substantial overhead and becomes prohibitive for long videos.

In our attention-graph formulation, the query-to-vision interaction is characterized by an evidence injection distribution $\boldsymbol{\pi}^{(0)}$, which measures the probability mass assigned by the query to each visual patch via cross-modal attention (see Appx. A.3). 
We use a lightweight query-to-patch relevance score $\mathbf{s}_{\text{evi}}$ as an efficient estimator of $\boldsymbol{\pi}^{(0)}$, following the spirit of Eq.~\ref{eq:evidence_localization} but on patch features:
\begin{equation}
\label{eq:obj_evi}
 \mathbf{s}_{\text{evi}}=
 \frac{1}{N} \cdot \hat{V}_{\text{patch}} \hat{Q}^\top \in \mathbb{R}^{T \times P}.
\end{equation}
$\mathbf{s}_{\text{evi}}$ assigns higher scores to patches that are more likely to be attended by the query, serving as candidate evidence for grounding.
It approximates query-to-vision evidence injection using simple feature similarity, avoiding the overhead of materializing full cross-attention maps.

However, directly taking the top-$k$ tokens is prone to redundancy.
Since high-scoring patches often cluster around the same object across nearby spatial locations, leading to near-duplicate selections that waste budget and reduce semantic coverage.
We therefore adopt Maximal Marginal Relevance (MMR) to balance \emph{query relevance} and \emph{non-redundancy}: 
at each step, it selects the candidate patch $p^{\star(t)}_{\text{obj}}$ for frame $t$ that maximizes
\begin{equation}
\label{eq:sti_mmr}
p^{\star(t)}_{\text{obj}}=\arg\max_{p\in\mathcal{C}}\Big[\lambda_{\text{mmr}} \cdot \mathbf{s}_{\text{evi}}^{(t,p)}-(1-\lambda_{\text{mmr}}) \cdot\max_{p'\in\mathcal{S}}(\hat{V}_{\text{patch}}^{(t,p)}\hat{V}_{\text{patch}}^{(t,p')\top})\Big],
\end{equation}
where $\mathcal{C}$ is the candidate patch index set at $t$, $\mathcal{S}$ is the already selected patch index set, $\hat{V}_{\text{patch}}^{(t,p)}$ is the normed feature of the patch $p$ at frame $t$, and $\lambda_{\text{mmr}}\in[0,1]$ controls the trade-off.
MMR selects query-relevant patches while penalizing candidates that are too similar to already selected ones. This encourages diverse object evidence and avoids wasting limited budget on near-duplicate patches around the same region.
The detailed efficient algorithm is recorded in Appx. B.

\subsection{Motion Token}
Object evidence is necessary but not always sufficient for VTG.
Precise temporal grounding requires capturing the state transitions and maintaining a traceable path that links evidence across frames.
To facilitate cross-frame routing, SemVID introduces motion tokens as explicit relays to capture state changes.

In dot-product attention, a token routes less mass to tokens with low feature similarity.
Consequently, long-range evidence propagation is most likely to bottleneck at frames where the visual state changes sharply.
We therefore identify motion tokens from regions with strong temporal feature variation by scoring each patch with a cheap token-level difference $m_{\text{mot}}^{(t,p)}$:
\begin{equation}
\label{eq:mot_diff}
 m_{\text{mot}}^{(t,p)}= \begin{cases}
 \|\hat V_{\text{patch}}^{(t+1,p)}-\hat V_{\text{patch}}^{(t,p)}\|_2, & t=1,\\[6pt]
 \|\hat V_{\text{patch}}^{(t,p)}-\hat V_{\text{patch}}^{(t-1,p)}\|_2, & t=T,\\[6pt]
 \tfrac{1}{2}\big(\|\hat V_{\text{patch}}^{(t,p)}-\hat V_{\text{patch}}^{(t-1,p)}\|_2+\|\hat V_{\text{patch}}^{(t+1,p)}-\hat V_{\text{patch}}^{(t,p)}\|_2\big), & \text{otherwise}.
 \end{cases}
\end{equation}
Large $m_{\text{mot}}^{(t,p)}$ indicates a strong local transition, where temporal connectivity is typically most fragile.
Identifying such regions helps preserve relay tokens that bridge evidence before and after the transition.
A formal definition of temporal connectivity is provided in Appx. A.3.

Since motion alone may be dominated by camera jitter or background changes, we make motion selection query-aware by fusing motion with query relevance:
\begin{equation}
\label{eq:mot_score}
 s_{\text{mot}}^{(t,p)}=(1-\beta)\, m_{\text{mot}}^{(t,p)} + \beta\, \max_{\hat{q}\in \hat{Q}}\bigl(\hat{V}_{\text{patch}} \hat{q}^\top\bigr),
\end{equation}
Given that irrelevant background motions usually have low query relevance,
we keep the top-$k^{(t)}_{\text{mot}}$ tokens according to $s_{\text{mot}}^{(t,p)}$, which makes motion selection query-aware by combining temporal change with query relevance. 
As a result, SemVID can retain meaningful transitions related to the query while suppressing irrelevant background or camera-induced motion.

\subsection{Context Token}
Without context anchors, pruning can create token-empty gaps or overly query-centric evidence, both of which distort temporal reasoning.
To keep each frame interpretable, we retain a small set of query-agnostic context tokens.

First, we select a \emph{proto} token $p^{\star}_{\text{proto}}$ that best represents the frame background by matching the frame-global mean feature:
\begin{equation}
\label{eq:sti_ctx}
p^{\star(t)}_{\text{proto}}=\arg\max_{p\in\{1,\dots,P\}} \hat{V}_{\text{patch}}^{(t,p)}\hat{V}_{\text{glb}}^{(t)\top}.
\end{equation}

Then, to complete the context budget $p^{\star(t)}_{\text{ctx}}$, we select the remaining $k_{\text{ctx}}-1$ tokens by a lightweight saliency score $s_{\text{sal}}^{(t,p)}=\left\|V_{\text{patch}}^{(t,p)}\right\|_2$, yielding:
\begin{equation}
\label{eq:sti_s_saliency}
\mathcal{P}^{(t)}_{\text{ctx}}=
\left\{p^{\star(t)}_{\text{proto}}\right\}
\cup
\operatorname{TopK}\!\left(\left\{s_{\text{sal}}^{(t,p)}\right\}_{p=1}^{P},\, k_{\text{ctx}}-1\right),
\end{equation}
where $\operatorname{TopK}(\cdot,k)$ returns the indices of the $k$ largest scores.
Although we keep only a few context tokens per frame, they are vital for perceiving coherent context and scene changes, avoiding fragmented reasoning.

\subsection{Evaluation Metrics}
\label{sec:exp:eval_metrics}
Pruning removes nodes and edges in the attention graph, which can degrade both evidence preservation and propagation.
We therefore quantify the quality of pruned graph with Evidence Retention (ER) and Connectivity Strength (CS).

\textbf{Evidence Retention.}
An effective pruning method should not remove the patches that carry query-critical evidence, so that query could still reach the same visual evidence as in the full video.
To measure this, we compare the evidence landing map before pruning, $\boldsymbol{\pi}^{(1)}_{\text{full}}$, with the one after pruning, $\boldsymbol{\pi}^{(1)}$, where $\boldsymbol{\pi}^{(1)}$ denotes the distribution of visual patches that the query finally routes to in the attention graph (Appx. A.3).
A direct cross-entropy comparison is problematic because pruned tokens have zero probability after pruning, which can make the loss infinite.
We therefore first compute the amount of original evidence mass that is still kept by the retained token set:
$\rho \;=\; \sum_{v\in\mathcal V'} \boldsymbol{\pi}_{\text{full}}^{(1)}(v)$.
Here, $\rho$ measures how much query-induced evidence from the full video remains after pruning.
We then use $\rho$ to build a smoothed post-pruning distribution $\bar{\boldsymbol{\pi}}^{(1)}$ and define ER as
\begin{equation}
\label{eq:er_ce}
\mathrm{ER}(\mathcal V')
=
\exp\big(\sum_{v\in\mathcal V} \boldsymbol{\pi}_{\text{full}}^{(1)}(v)\log \bar{\boldsymbol{\pi}}^{(1)}(v)\big),
\quad
\bar{\boldsymbol{\pi}}^{(1)}(v)=
\begin{cases}
\rho\cdot \boldsymbol{\pi}^{(1)}(v), & v\in\mathcal V',\\[6pt]
\dfrac{1-\rho}{|\mathcal V\setminus\mathcal V'|}, & v\notin\mathcal V',
\end{cases}
\end{equation}
where $\mathcal V'\subseteq\{V_{\text{patch}}^{(1,1)},\dots,V_{\text{patch}}^{(T,P)}\}$ is the retained token set.
We exponentiate the negative cross-entropy to bound ER in $(0,1]$.
If pruning removes query-critical tokens, $\pi^{(1)}$ will deviate from the full one, leading to lower ER.

\textbf{Connectivity Strength.}
To quantify whether pruning preserves a traceable evidence chain, we measure how much attention mass can be routed across adjacent frames, so that information could move smoothly from one frame to the next.
At layer $\ell$, we define the cross-frame routing mass from frame $t$ to $t{+}1$ as
$\Gamma^{(\ell)}_{\mathcal V'}(t)=\sum_{u_i\in\mathcal{V}'_t}\sum_{u_j\in\mathcal{V}'_{t+1}}\mathbb{P}^{(\ell)}(u_i\!\to\!u_j)$,
where $\mathcal{V}'_t$ denotes the retained tokens in frame $t$ and $\mathbb{P}^{(\ell)}(u_i\!\to\!u_j)$ is the layer-$\ell$ routing probability (i.e., the attention-based transition mass) from token $u_i$ to token $u_j$ in the attention graph.
The detailed implementation of $\mathbb{P}^{(\ell)}(u_i\!\to\!u_j)$ is provided in Appx. A.2.
We then aggregate $\Gamma^{(\ell)}_{\mathcal V'}(t)$ over time and layers to obtain CS:
\begin{equation}
\label{eq:cfc_metric}
\mathrm{CS}(\mathcal V')=\frac{1}{L}\sum_{\ell=1}^{L}\sum_{t=1}^{T-1}\Gamma_{\mathcal V'}^{(\ell)}(t),
\end{equation}
where $L$ is the layer amount.
Intuitively, larger $\Gamma^{(\ell)}_{\mathcal V'}(t)$ means that the retained tokens form stronger temporal links across frames.
In other words, the pruned video still provides a connected path for multi-hop evidence propagation, which helps the model localize temporal boundaries more reliably.
\section{Experiments}

\subsection{Experimental Setting}
\textbf{Benchmarks.}
We evaluate SemVID on Video Temporal Grounding (VTG) in two standard long-video grounding benchmarks, Charades-STA \cite{sigurdsson2016hollywood} and ActivityNet-Grounding \cite{caba2015activitynet}.
In the appendix, we also evaluate VideoQA on Video-MME \cite{fu2025video} and LongVideoBench \cite{wu2024longvideobench}.
The details of benchmarks are recorded in Appx. C.

\noindent
\textbf{Metrics.}
Following~\cite{ren2024timechat,zeng2024timesuite}, we report mean Intersection over Union (mIoU) and $R1$ at tIoU thresholds $m\!\in\!\{0.3,0.5,0.7\}$.
TFLOPs are estimated for efficiency.

\noindent
\textbf{Implementation Details.}
We set up a unified evaluation protocol for fair comparisons detailed in Appx. C.
SemVID is evaluated on Qwen3-VL-4B/8B-Thinking \cite{yang2025qwen3}, Qwen2.5-VL-7B-Instruct \cite{bai2025qwen2}, and LLaVA-OneVision-7B \cite{li2024llava}.
To our knowledge, this is the first study of pruning on Qwen3-VL.
We also adapt FastVID \cite{shen2025fastvid} and VisionZip \cite{yang2025visionzip} to Qwen3-VL as baselines.
Unless noted, we set $\alpha=0.6$ (Eq. \ref{eq:ba_w}), $\lambda_{\text{mmr}}=0.8$ (Eq. \ref{eq:sti_mmr}), $\beta=0.5$ (Eq. \ref{eq:mot_score}), and $k_{\text{ctx}}=3$ (Eq. \ref{eq:sti_s_saliency}).

\subsection{Comparisons with State-of-the-Art Methods}
\subsubsection{Qwen3-VL.}

\cref{tab:main_qw3} shows that our Qwen3-VL-based SemVID consistently achieves the best accuracy-efficiency trade-off across model sizes and budgets.
Under an extremely aggressive 12.5\% budget, SemVID retains up to 95.4\% of the original mIoU while largely preserving performance at high IoU ($R1@0.7$), indicating precise boundary localization rather than coarse retrieval.

\begin{table*}[!hb]
\centering
\small
\setlength{\tabcolsep}{3pt}
\begin{adjustbox}{width=1\textwidth,center}
\begin{tabular}{l| cccc c| cc| cccc c| cc}
\toprule
\multicolumn{1}{l|}{} &
\multicolumn{7}{c|}{ActivityNet-Grounding} &
\multicolumn{7}{c}{Charades-STA} \\
\cmidrule(lr){2-8}\cmidrule(lr){9-15}
Method &R1@0.3 & R1@0.5 & R1@0.7 & mIoU & (\%)& ER& CS &
R1@0.3 & R1@0.5 & R1@0.7 & mIoU & (\%) & ER& CS \\
\midrule
\rowcolor{lightgray}
\multicolumn{15}{l}{\textbf{Qwen3-VL-4B, Retain 12.5\% Tokens}} \\
\midrule
Qwen3-VL-4B    &   
54.60 & 37.54 & 23.74 & 40.33 & 100 & $1e^{4}$ & 64.4 &
79.62 & 65.99 & 40.81 & 56.07 & 100 & $1e^{4}$ & 81.5 \\

VisionZip \cite{yang2025visionzip} &
26.02 & 15.85 & 9.58 & 19.89 & 49.3 & 1.6 & 22.4 &
56.91 & 37.20 & 16.72 & 36.83 & 65.7 & 3.0 & 26.4 \\

FastVID \cite{shen2025fastvid}     &
45.70 & 29.32 & 17.30 & 33.16 & 82.2 & 3.0 & \textbf{39.7} &
55.33 & 33.32 & 15.47 & 35.98 & 64.2 & 2.9 & 29.8 \\

\textbf{SemVID (ours)} &
\textbf{51.96} & \textbf{34.73} & \textbf{22.18} & \textbf{38.49} & \textbf{95.4} & \textbf{4.5} & 31.6 &
\textbf{74.17} & \textbf{56.91} & \textbf{30.67} & \textbf{49.89} & \textbf{89.0} & \textbf{5.6} & \textbf{33.5} \\

\midrule
\rowcolor{lightgray}
\multicolumn{15}{l}{\textbf{Qwen3-VL-4B, Retain 25\% Tokens}} \\
\midrule
Qwen3-VL-4B    &
54.60 & 37.54 & 23.74 & 40.33 & 100 & $1e^{4}$ & 64.4 &
79.62 & 65.99 & 40.81 & 56.07 & 100 & $1e^{4}$ & 81.5 \\

VisionZip \cite{yang2025visionzip} &
30.74 & 19.62 & 12.15 & 23.30 & 57.8 & 2.8 & 30.8 &
64.11 & 47.28 & 23.79 & 43.09 & 76.9 & 3.6 & 47.3 \\

FastVID \cite{shen2025fastvid}     &
46.54 & 30.24 & 18.42 & 34.33 & 85.1 & 3.8 & \textbf{57.3} &
57.93 & 39.63 & 19.52 & 38.13 & 68.0 & 3.1 & 53.7\\

\textbf{SemVID (ours)} &
\textbf{52.84} & \textbf{35.68} & \textbf{22.51} & \textbf{39.06} & \textbf{96.9} & \textbf{6.5} & 55.2 &
\textbf{76.91} & \textbf{61.08} & \textbf{33.17} & \textbf{52.31} & \textbf{93.3} & \textbf{9.8} & \textbf{56.1} \\

\midrule
\rowcolor{lightgray}
\multicolumn{15}{l}{\textbf{Qwen3-VL-8B, Retain 12.5\% Tokens}} \\
\midrule
Qwen3-VL-8B  &
52.86 & 35.81 & 22.86 & 39.10 & 100 & $1e^{4}$ & 94.7 &
80.03 & 66.59 & 42.20 & 56.67 & 100 & $1e^{4}$ & 121.5 \\

VisionZip \cite{yang2025visionzip} & 
10.43 & 6.18 & 3.91 & 8.12 & 20.8 & 1.0 & 14.3 &
21.43 & 15.73 & 7.69 & 14.52 & 25.6 & 1.1 & 17.7  \\

FastVID \cite{shen2025fastvid}     &
40.53 & 26.16 & 15.91 & 30.61 & 78.3 & 3.5 & 28.4 &
52.84 & 35.58 & 16.36 & 35.26 & 62.2 & 2.8 & 26.7\\

\textbf{SemVID (ours)} &
\textbf{48.32} & \textbf{31.93} & \textbf{20.47} & \textbf{36.21} & \textbf{92.6} & \textbf{6.8} & \textbf{29.0} &
\textbf{73.45} & \textbf{56.24} & \textbf{31.11}& \textbf{49.93} & \textbf{88.1} & \textbf{5.7} & \textbf{30.5} \\

\midrule
\rowcolor{lightgray}
\multicolumn{15}{l}{\textbf{Qwen3-VL-8B, Retain 25\% Tokens}} \\
\midrule
Qwen3-VL-8B  &
52.86 & 35.81 & 22.86 & 39.10 & 100 & $1e^{4}$ & 94.7 &
80.03 & 66.59 & 42.20 & 56.67 & 100 & $1e^{4}$ & 121.5 \\

VisionZip \cite{yang2025visionzip} &
14.24 & 8.95 & 5.63 & 10.92 & 27.9 & 1.8 & 20.1&
36.13 & 28.31 & 15.86 & 24.96 & 44.4 & 2.9 & 21.8 \\

FastVID \cite{shen2025fastvid}     &
41.73 & 27.23 & 16.93 & 31.63 & 80.9 & 3.9 & 45.8&
55.38 & 38.58 & 19.49 & 37.12 & 65.5 & 3.5 & 24.2 \\

\textbf{SemVID (ours)} & 
\textbf{50.30} & \textbf{33.49} & \textbf{21.50} & \textbf{37.54} & \textbf{96.0} & \textbf{8.8} & \textbf{51.2} &
\textbf{76.30} & \textbf{60.11} & \textbf{34.55} & \textbf{52.66} & \textbf{93.0} & \textbf{9.5} & \textbf{51.6} \\

\bottomrule
\end{tabular}
\end{adjustbox}
\caption{VTG results with Qwen3-VL under 12.5\% and 25\% visual token retention. Percentages are relative to the original. Evidence Retention (ER) and Evidence Retention (CS) are attention graph metrics introduced in \cref{sec:exp:eval_metrics}. The unit for ER is $1e^{-4}$.}
\label{tab:main_qw3}
\end{table*}

A key observation is that the gap between methods is well explained by our two attention-graph diagnostics: Evidence Retention (ER) and Connectivity Strength (CS).
VisionZip, a redundancy-driven approach, tends to over-merge temporally adjacent visual states, which both suppresses boundary-critical evidence for VTG and removes intermediate relay tokens, leading to pronounced degradations in ER and CS. 
In contrast, FastVID is saliency-driven and better preserves graph connectivity, but it concentrates the limited budget on a small set of anchor frames, leaving boundary-adjacent evidence sparsely represented and reducing effective ER. 
SemVID, instead, explicitly allocates tokens to both query-relevant evidence and high-variation transition frames, thereby constructing a comprehensive and temporally continuous evidence chain that jointly sustains ER and CS under aggressive pruning.

At 25\% retention, SemVID approaches near-lossless compression on both benchmarks, retaining up to 96.9\% of the original mIoU.
This strong retention further validates that preserving both ER and CS is crucial for VTG.

\subsubsection{Qwen2.5-VL.}

\begin{table*}[t]
\centering
\small
\setlength{\tabcolsep}{3pt}
\begin{adjustbox}{width=1\textwidth,center}
\begin{tabular}{l| cccc c |cc| cccc c |cc}
\toprule
\multicolumn{1}{l|}{} &
\multicolumn{7}{c|}{ActivityNet-Grounding} &
\multicolumn{7}{c}{Charades-STA} \\
\cmidrule(lr){2-8}\cmidrule(lr){9-15}
Method &
R1@0.3 & R1@0.5 & R1@0.7 & mIoU & (\%) & ER& CS &
R1@0.3 & R1@0.5 & R1@0.7 & mIoU & (\%) & ER& CS  \\
\midrule
\rowcolor{lightgray}
\multicolumn{15}{l}{\textbf{Qwen2.5-VL-7B, Retain 12.5\% Tokens}} \\
\midrule
Qwen2.5-VL-7B     &    
29.22 & 15.77 & 7.49 & 22.25 & 100 & $1e^{4}$ & 71.8 &
77.39 & 59.33 & 33.82 & 56.85 & 100 & $1e^{4}$ & 90.9 \\

FastV \cite{fastv} &
\multicolumn{5}{c|}{Out-of-Memory (OOM)} & -& - 
&\multicolumn{5}{c|}{Out-of-Memory (OOM)} & - & -\\
VScan \cite{zhang2025vscan} &
\multicolumn{5}{c|}{Out-of-Memory (OOM)} & - & - 
&\multicolumn{5}{c|}{Out-of-Memory (OOM)} & - & - \\

DART \cite{dart}          & 
16.04 & 8.24 & 4.06 & 12.38 & 55.6 & 2.7 & 14.3 &
35.65 & 25.19 & 12.85 & 24.30 & 42.7 & 2.8 & 15.7 \\

ToME \cite{bolya2022tome} & 
20.22 & 10.24 & 4.44 & 15.86 & 71.3 & 3.8 & 18.3 &
43.46 & 29.08 & 14.62 & 29.93 & 52.6 & 3.3 & 20.7 \\

TokenSculpt \cite{qatokensculpt} & 
20.67 & 10.54 & 4.73 & 16.20 & 72.8 & 4.1& 18.8 &
44.19 & 29.19 & 14.81 & 30.08 & 52.9 & 3.6 & 21.4 \\

FastVID \cite{shen2025fastvid}     &
20.89 & 10.53 & 4.66 & 16.28 & 73.1 & 3.9 & 22.7 &
44.08 & 29.96 & 15.68 & 30.57 & 53.8 & 3.7 & 21.9  \\

VisionZip \cite{yang2025visionzip} & 
20.87 & 10.51 &4.65 & 16.33 & 73.3 & 4.3 & 18.9 &
50.13 & 33.27 & 15.88 & 33.51 & 58.9 & 4.1 & 20.1 \\

\textbf{SemVID (ours)} & 
\textbf{21.74} & \textbf{10.69} & \textbf{4.63} & \textbf{17.21} & \textbf{77.4} & \textbf{4.5 }& \textbf{23.1} &
\textbf{54.01} & \textbf{35.67} & \textbf{18.23} & \textbf{36.54} & \textbf{64.3} & \textbf{4.2} & \textbf{24.9} \\

\bottomrule
\end{tabular}
\end{adjustbox}
\caption{VTG results with Qwen2.5-VL under 12.5\% retention. FastV and VScan materialize full self-attention matrix and lead to OOM in long videos.}
\label{tab:main_qw2}
\end{table*}

We further evaluate SemVID on Qwen2.5-VL-7B under a 12.5\% visual-token budget in \cref{tab:main_qw2}. 
SemVID again achieves the best overall performance on both Charades-STA and ActivityNet, and the results suggest that VTG accuracy strongly correlates with jointly high ER and CS.
Existing baselines largely overlook temporal connectivity, whereas SemVID explicitly optimizes it and improves CS by 23.9\% over VisionZip (20.1 to 24.9).
Importantly, several query-driven baselines are not deployable on long videos because they require materializing full cross-attention weights, leading to out-of-memory errors. 
SemVID avoids this overhead by relying on lightweight similarity and temporal-difference proxies, making it practical for long-video inference.

\begin{figure}[t]
  \centering
    \begin{minipage}[t]{0.46\linewidth}
    \centering
    \includegraphics[width=\linewidth]{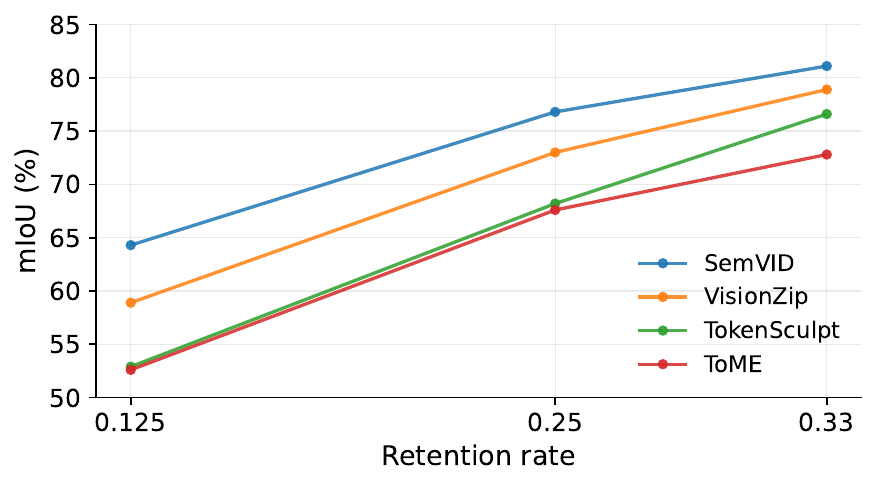}
    \caption{mIoUs on Charades-STA under different token retention ratios. }
    \label{fig:retention_tendency}
  \end{minipage}\hfill
  \begin{minipage}[t]{0.46\linewidth}
    \centering
    \includegraphics[width=\linewidth]{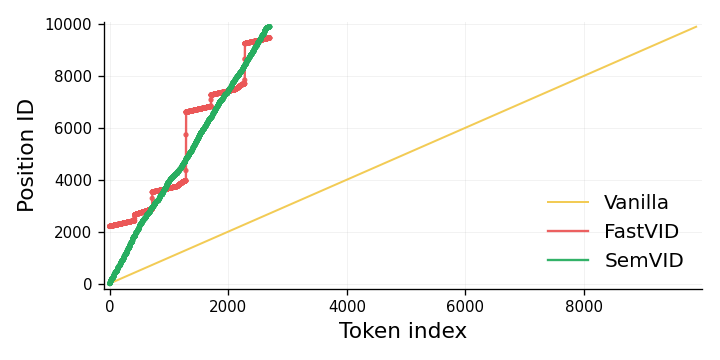}
    \caption{Position-ID trajectories before and after pruning. Pruning alters the trajectory relative to the original.}
    \label{fig:pos_emb}
  \end{minipage}
\end{figure}

We additionally vary the retention ratio and plot the mIoU trends in \cref{fig:retention_tendency}.
SemVID shows a markedly slower degradation under aggressive pruning, indicating that semantic evidence allocation is particularly effective when the token budget is the bottleneck under the realistic regime for long videos.

Furthermore, we observe that Qwen2.5-VL suffers a larger pruning-induced performance drop than Qwen3-VL.
A plausible reason is the positional encoding.
Qwen2.5-VL relies on RoPE \cite{su2024roformer}, and irregular token retention can distort the position-ID trajectory (\cref{fig:pos_emb}), introducing and amplifying bias in timestamp perception \cite{endo2025feather,ye2025atp}.
In contrast, Qwen3-VL uses explicit timestamp tokens before each frame, making temporal cues more robust to pruning.

\subsection{Ablation Study}

\subsubsection{Budget Allocation (BA).}
\begin{table*}[t]
\centering
\small
\setlength{\tabcolsep}{3pt}
\begin{adjustbox}{width=0.97\textwidth,center}
\begin{tabular}{l l cc | cccc c |cc}
\toprule
\multicolumn{4}{l|}{} &
\multicolumn{7}{c}{Charades-STA} \\
\cmidrule(lr){5-11}
  & Method & SBA & STI & R1@0.3 & R1@0.5 & R1@0.7 & mIoU & (\%) & ER & CS\\
\midrule
\multirow{7}{*}{(a)} & Qwen3-VL-4B         & -& -&
79.62 & 65.99 & 40.81 & 56.07 & 100 & $1e^{4}$ & 81.5 \\

                     & Sampling with Fixed-Interval     &  &  &
61.18 & 43.60 & 20.16 & 39.80 & 71.0 & 2.8 & 32.0 \\

                     & Random Sampling        &  &  &
63.98 & 44.67 & 21.10 & 41.76 & 74.4 & 3.0 & 27.7 \\

                     & Sampling with Query-Relevance     &  &  &
67.31 & 49.56 & 25.88 & 44.97 & 80.2 & 5.5 & 20.6 \\

                     & Semantic Budget, Random Selection    & \checkmark&  &
66.25 & 47.68 & 23.79 & 43.26 & 77.2 & 3.6 & 28.1 \\

                     & Uniform Budget, Semantic Selection        & & \checkmark&
73.01 & 54.91 & 29.60 & 48.50 & 86.5 & 5.1& 33.2\\

                     & \textbf{Full, Semantic Budget and Selection}  & \checkmark & \checkmark&
\textbf{74.17} & \textbf{56.91} & \textbf{30.67} & \textbf{49.89} & \textbf{89.0} & \textbf{5.6}& \textbf{33.5}\\

\midrule

\multirow{3}{*}{(b)} & Qwen3-VL-4B        & -& -&
79.62 & 65.99 & 40.81 & 56.07 & 100 & $1e^{4}$ & 81.5 \\

                     & FastVID \cite{shen2025fastvid}        &  &  &
55.33 & 33.32 & 15.47 & 35.98 & 64.2 & 2.9 & \textbf{29.8} \\

                     & \textbf{FastVID + Our Semantic Budget}     & \checkmark&  &
\textbf{73.69} & \textbf{55.57} & \textbf{29.79} & \textbf{48.88} & \textbf{87.2} & \textbf{5.3}& 26.4\\
\bottomrule

\end{tabular}
\end{adjustbox}
\caption{Ablation results on Semantic Budget Allocation (SBA) and Semantic Token Identification (STI). (a) Comparison with different sampling strategies. (b) Implementing our budget allocation module into FastVID.}
\label{tab:ablation_ba}
\end{table*}
Comparing \emph{Random Sampling} with \emph{Semantic Budget, Random Selection} in Table~\ref{tab:ablation_ba}(a), BA improves the retained mIoU by 2.8\% and boosts ER from 3.0 to 3.6.
Since within-frame token selection remains random, this gain isolates the contribution of frame-level budget allocation.
Specifically, BA introduces a lightweight prior to estimate per-frame information density, thereby increasing the likelihood of retaining query-relevant evidence and improving ER. 
Moreover, it prevents the budget from collapsing onto a few salient frames by allocating tokens to boundary-adjacent and transition moments, which helps preserve temporal connectivity and thus maintains CS.

To test generality and modularity, we plug our BA into FastVID in Table~\ref{tab:ablation_ba}(b), which substantially improves retained mIoU from 64.2\% to 87.2\%. 
As shown in Appx. D.5, FastVID relies on sparse anchor frames, which over-compresses boundary-adjacent moments, blurs transition cues, and fragments the temporal evidence chain. 
Our BA reallocates tokens toward query-relevant and high-variation transition frames and thus improving boundary evidence coverage.

\subsubsection{Semantic Token Identification (STI).}
Table~\ref{tab:ablation_ba}(a) shows that under a uniform per-frame budget, replacing fixed-interval sampling with STI improves the retained mIoU by 8.7\%.
This indicates that once temporal coverage is fixed, token selection becomes the dominant factor: STI prioritizes query-aligned evidence while avoiding redundant selections, thereby significantly increasing ER without collapsing CS.
Combining BA and STI yields the best results, outperforming either component alone and retaining 88.1\% of the original performance.
It demonstrates that pruning for VTG requires both BA to maintain frame coverage and evidence-aware STI to preserve diverse evidence and traceable chains.

\begin{table}[t]
\centering
\small

\begin{minipage}[t]{0.49\textwidth}
\centering
\setlength{\tabcolsep}{3pt}
\begin{adjustbox}{width=\linewidth,center}
\begin{tabular}{l l| c c| cc}
\toprule
\multicolumn{2}{l|}{} &
\multicolumn{4}{c}{Charades-STA} \\
\cmidrule(lr){3-6}
  & Method & mIoU & (\%) & ER & CS\\
\midrule
                    & Qwen3-VL-4B        &
56.07 & 100 & $1e^{4}$ & 81.5  \\
\midrule

\multirow{2}{*}{(a)}   & Attention Selection    &
46.57 & 83.1 & 5.0& 32.8\\

                       & \textbf{Relevance Selection}    &
\textbf{49.89} & \textbf{89.0} & \textbf{5.6}& \textbf{33.5}\\
\midrule

\multirow{3}{*}{(b)}  & $\lambda_{\text{mmr}}=1$ (w/o MMR)    &
49.44 & 88.2 & 5.4& 33.3\\

                     & $\mathbf{\lambda_{\text{mmr}}=0.8}$   &
\textbf{49.89} & \textbf{89.0}& \textbf{5.6}& 33.5\\

                     & $\lambda_{\text{mmr}}=0.6$           &
49.25 & 87.8 & 5.3& \textbf{33.9}\\
\midrule

\multirow{6}{*}{(c)}  & $\alpha=0$ (w/o Obj. Token)    &
49.48 & 88.2 & 5.4& \textbf{33.9}\\

                     & $\alpha=0.2$                          &
49.56 & 88.4& 5.5& 33.7\\

                     & $\alpha=0.4$                                      &
49.79 & 88.8 & 5.6& 33.6\\

                     & $\mathbf{\alpha=0.6}$      &
\textbf{49.89} & \textbf{89.0}& 5.6& 33.5\\

                     & $\alpha=0.8$        &
49.56 & 88.4& 5.6& 30.7\\

                     & $\alpha=1.0$ (w/o Mot. Token)       &
48.11 & 85.8 & \textbf{5.7}& 23.3\\

\bottomrule
\end{tabular}
\end{adjustbox}
\caption{Ablation on object tokens. 
(a) Different selection strategies. 
(b) Effectiveness of MMR (Eq. \ref{eq:sti_mmr}). 
(c) Different object-motion token ratios (Eq. \ref{eq:ba_w}). 
}
\label{tab:ablation_obj_sti}
\end{minipage}\hfill
\begin{minipage}[t]{0.49\textwidth}
\centering
\setlength{\tabcolsep}{3pt}
\begin{adjustbox}{width=\linewidth,center}
\begin{tabular}{l l| c c| cc}
\toprule
\multicolumn{2}{l|}{} &
\multicolumn{4}{c}{Charades-STA} \\
\cmidrule(lr){3-6}
  & Method & mIoU & (\%) & ER & CS\\
\midrule
                    & Qwen3-VL-4B        &
56.07 & 100 &  $1e^{4}$ & 81.5 \\
\midrule

\multirow{2}{*}{(a)} & $\beta=0$ (w/o query-aware)        &
49.04& 87.5& 5.5& 32.1\\

                     & $\mathbf{\beta=0.5}$     &
\textbf{49.89} & \textbf{89.0} & \textbf{5.6}& \textbf{33.5}\\
\midrule

\multirow{4}{*}{(b)} & $k_{\text{ctx}}=0$ (w/o Ctx. Token)        &
49.20 & 87.7 & \textbf{5.6} & 31.9\\

                     & $k_{\text{ctx}}=1$ (proto only)    &
49.35 & 88.0 & \textbf{5.6} & 32.4\\

                     & $\mathbf{k_{\text{ctx}}=3}$     &
\textbf{49.89} & \textbf{89.0} & \textbf{5.6}& 33.5\\

                     & $k_{\text{ctx}}=10$     &
48.52 & 86.5 & 4.8& \textbf{34.6}\\

\bottomrule
\end{tabular}
\end{adjustbox}
\caption{Ablation on motion and context token selection.
(a) Different query-aware weights in background change suppression (Eq. \ref{eq:mot_score}).
(b) Different context token selection strategies ($k_{\text{ctx}}$ in Eq. \ref{eq:sti_s_saliency} and the proto token in Eq. \ref{eq:sti_ctx}). 
}
\label{tab:ablation_mot_ctx_selection}
\end{minipage}
\end{table}


\textbf{Object Tokens.}
Table~\ref{tab:ablation_obj_sti}(a) compares two evidence selection strategies. \textit{Attention selection} uses the raw query and key projection matrix from the model’s first attention block to compute a lightweight cross-attention between the language and visual patches.
However, attention sinks can bias the weights toward non-semantic patches, hindering accurate identification of right evidence.
In contrast, using direct query relevance is both more efficient and more precise for evidence localization, resulting in a 5.9\% gain in retained performance.
Further, Table~\ref{tab:ablation_obj_sti}(b) shows that Maximal Marginal Relevance (MMR) expands the evidence set to cover complementary object parts and nearby contextual cues, yielding a denser and more stable evidence extraction, as reflected by higher ER.

\textbf{Motion Tokens.}
Ablations on the object-motion ratio in Table~\ref{tab:ablation_obj_sti}(c) show that removing motion tokens causes a clear drop in both mIoU and CS.
This supports our claim that VTG is limited not only by the presence of evidence but also by their connectivity.
Our motion tokens serve as relay nodes that strengthen CS, enabling the model to connect evidence via multi-hop attention. 
Although allocating motion tokens consumes a small fraction of the budget, it is crucial for maintaining a coherent evidence chain, while the majority of tokens should remain dedicated to object tokens to maximize evidence coverage.

Additionally, Table~\ref{tab:ablation_mot_ctx_selection}(a) shows that query-aware filtering improves CS by suppressing background transitions and focusing on query-relevant state changes.

\textbf{Context Tokens.}
Varying $k_{\text{ctx}}$ in Table~\ref{tab:ablation_mot_ctx_selection}(b) shows that a small number of context tokens is necessary.
Completely removing context anchors hurts evidence connectivity, while over-allocating context dilutes boundary-critical evidence.
This highlights the intended role of context tokens: that they are not competing evidence but lightweight anchors that capture global scene context and stabilize temporal connectivity under aggressive pruning.

\subsection{Inference Latency}
Table~\ref{tab:efficiency} reports an efficiency comparison. 
SemVID achieves the best accuracy-efficiency balance: under the same 12.5\% token budget, it substantially outperforms FastVID and VisionZip in accuracy.
In terms of latency, SemVID reduces the prefill time from 1263.4\,ms to 217.7\,ms, yielding a 5.8$\times$ speedup over the original.
Although FastVID shows a slightly lower pruning overhead, this difference is negligible compared to the dominant LLM forward cost.
Overall, SemVID provides near-FastVID efficiency while preserving markedly stronger performance, making it a practical choice when both throughput and quality matter.

\begin{table*}[t]
\centering
\small
\setlength{\tabcolsep}{3pt}
\begin{adjustbox}{width=0.88\textwidth,center}
\begin{tabular}{l | cc| cc| cccc | cc}
\toprule
\multicolumn{1}{l|}{} &
\multicolumn{2}{c|}{Tokens} &
\multicolumn{2}{c|}{TFLOPs} &
\multicolumn{4}{c|}{Prefill Time} &
\multicolumn{2}{c}{ActivityNet} \\
\cmidrule(lr){2-3}\cmidrule(lr){4-5}\cmidrule(lr){6-9}\cmidrule(lr){10-11}
Method  & \# & (\%) & Value & (\%) &
Pruning & LLM Forward & Total & Speed & mIoU & (\%) \\
\midrule
Qwen3-VL-4B         & 10460 & 100 & 59.4 & 100 &
- & 1263.4 & 1263.4 & 1$\times$ & 40.33 & 100 \\

\midrule

VisionZip \cite{yang2025visionzip}  & 1307 & 12.5 & 4.8 & 8.7 &
710.9 & 184.3 & 895.2 & 1.4$\times$ & 19.89 & 49.3 \\

FastVID \cite{shen2025fastvid}     & 1370 & 13.1 & 5.4 & 9.1 &
\textbf{23.8} & 185.9 & \textbf{209.7} & \textbf{6.0$\times$} & 33.16 & 82.2 \\

\textbf{SemVID}                    & 1307 & 12.5 & 4.8 & 8.7 &
33.1 & 184.6 & 217.7 & 5.8$\times$ & \textbf{38.49} & \textbf{95.4} \\

\bottomrule
\end{tabular}
\end{adjustbox}
\caption{Efficiency Comparison on Qwen3-VL. Prefill time refers to the latency to the first generated token, which comprises pruning latency introduced by token selection.}
\label{tab:efficiency}
\end{table*}

\begin{figure}[b]
    \centering
    \includegraphics[width=1\linewidth]{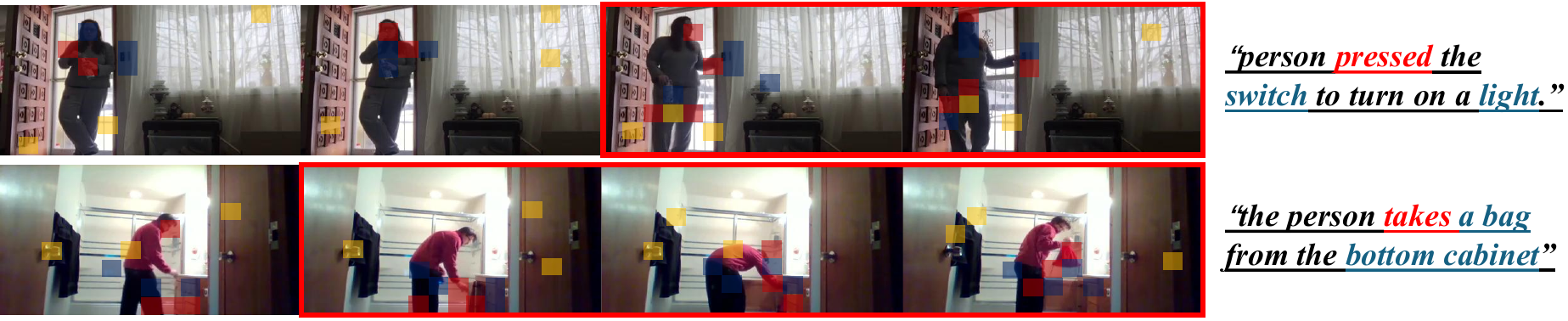}
    \caption{Visualization results. \textcolor{blue}{blue boxes} denote object tokens, \textcolor{red}{Red boxes} indicate motion tokens,  and \textcolor{yellow!30!orange}{yellow boxes} represent context tokens. }
    \label{fig:visualization}
\end{figure}

\subsection{Visualization}

\cref{fig:visualization} qualitatively illustrates SemVID's role-aware token selection.
SemVID retains noticeably more tokens near event boundaries, where state transitions occur and grounding evidence is most informative.
This aligns with the goal of preserving a coherent evidence chain rather than uniformly compressing the video.

We further observe distinct roles for different semantic tokens.
\emph{Object tokens} capture query-mentioned entities (e.g., switch, cabinet), treating them as evidence.
MMR (Eq.~\ref{eq:sti_mmr}) avoids redundant duplicates of a single object.
\emph{Motion tokens} concentrate on action regions and often highlight the decisive transition cues that bridge pre-boundary and post-boundary evidence.
\emph{Context tokens} cover stable background anchors that help maintain temporal coherence under aggressive pruning. 
These qualitative behaviors are consistent with the quantitative ER and CS improvements and the strong VTG performance.

Additional visualizations of the effects of our budget allocation and evidence retention are provided in Appx. D.

\subsection{Discussion}
While SemVID preserves the evidence chain with role-aware token selection, a limitation of our implementation lies in the coarse Budget Allocation (BA).
BA jointly considers query-frame relevance and inter-frame variation, where the latter is approximated by frame feature differences.
This proxy can be biased by camera or irrelevant motions and thus absorb part of the budget (Appx. D.5), potentially weakening coverage of subtle actions.
A related failure case is that background motion may produce strong local variation and interfere with motion-token selection.
In practice, these effects are largely mitigated by our per-frame context floor and query-aware motion filtering: the former prevents weak evidence from being completely pruned, while the latter suppresses query-irrelevant transitions.
Nevertheless, more robust motion identification remains an important future direction.
We report a dedicated analysis of sensitivity to motion amplitude and include additional evaluation on VideoQA in Appx. D.

\section{Conclusion}

We revisit training-free visual token pruning for Video Temporal Grounding (VTG) under an evidence chain formulation and formalize two VTG-tailored metrics, evidence retention and connectivity strength.
We propose SemVID, a plug-and-play training-free pruning framework that explicitly optimizes these two objectives.
SemVID performs semantic budget allocation and selects a role-aware token set: 
object tokens preserve diverse query-aligned evidence, motion tokens act as query-filtered transition relays, and lightweight context tokens stabilize scene continuity. 
Extensive experiments demonstrate that SemVID consistently delivers a strong accuracy-efficiency trade-off, retaining substantially higher performance over strong baselines while significantly accelerating prefill.
By explicitly preserving the right evidence and keeping it connected, SemVID provides a simple yet effective recipe for making long-video VTG practical.

%
%
\bibliographystyle{splncs04}
\bibliography{main}

\clearpage
\appendix

\section*{Supplementary Material}

\section{Preliminary}

\subsection{Attention Graph.}
\label{appx:pre:attention_graph}
To simplify the token-to-token interactions in a VLM, we decompose them into language-to-vision cross-attention and vision self-attention.
For each Transformer layer $\ell$, vision self-attention induces a directed weighted graph
$G^{(\ell)}=\bigl(\mathcal{V},\mathcal{E}^{(\ell)},w^{(\ell)}\bigr)$,
where $\mathcal{V}$ is the set of visual tokens, $\mathcal{E}^{(\ell)}$ contains directed token-to-token edges,
and $w^{(\ell)}$ assigns a nonnegative weight to each edge.
Then we compute the cross-attention from language tokens $\mathcal{Q}$ to visual tokens $\mathcal{V}$.
Following prior research \cite{abnar2020quantifying, carion2020end}, we use the weights in the $L$-th layer, where $L$ is the number of layers within a Transformer model, to induce an \textbf{injection distribution} $\boldsymbol{\pi}^{(L)}\in\mathbb{R}^{|\mathcal{V}|}$ over $\mathcal{V}$.
Intuitively, $\boldsymbol{\pi}^{(L)}$ tells where the query reads from the visual tokens, which is the starting point of evidence.

\subsection{Evidence Flow}
\label{appx:pre:evidence_flow}
Given the attention graph at layer $\ell$, $G^{(\ell)}=\bigl(\mathcal{V},\mathcal{E}^{(\ell)},w^{(\ell)}\bigr)$,
we define the Markov transition matrix $\mathbf{P}^{(\ell)}\in\mathbb{R}^{|\mathcal{V}|\times|\mathcal{V}|}$ whose entries are
\begin{equation}
\label{eq:markov_routing}
\mathbf{P}^{(\ell)}_{i,j}
=\mathbb{P}^{(\ell)}(u_i\!\to\! u_j)
=\frac{w^{(\ell,u_i\!\to\! u_j)}}
{\sum_{j':\, (u_i\to u_{j'})\in \mathcal{E}^{(\ell)}} w^{(\ell,u_i\!\to\! u_{j'})}} \, ,
\end{equation}
where $u_i,u_j\in \mathcal{V}$ are patch tokens and $(u_i\!\to\!u_j)\in\mathcal{E}^{(\ell)}$ denotes a directed self-attention edge from $u_i$ to $u_j$.
By construction, each row of $\mathbf{P}^{(\ell)}$ sums to one, hence $\mathbf{P}^{(\ell)}$ is row-stochastic and specifies how a token probabilistically routes information to its attended neighbors at layer $\ell$.

Let $\boldsymbol{\pi}^{(\ell)}$ denote the evidence distribution over visual tokens at layer $\ell$.
Starting from the injection distribution $\boldsymbol{\pi}^{(L)}$ induced by cross-attention, evidence is propagated backward through vision self-attention layers by
\begin{equation}
\label{eq:pi_layer}
\boldsymbol{\pi}^{(\ell)}=\mathbf{P}^{(\ell+1)\top}\boldsymbol{\pi}^{(\ell+1)}\in\mathbb{R}^{|\mathcal{V}|},\qquad \ell=L-1,\dots,1.
\end{equation}
The resulting $\boldsymbol{\pi}^{(1)}$ is the \textbf{landing distribution} of evidence flow over the input patch tokens.
A larger $\pi^{(1)}(v)$ means that token $v$ receives more query-induced evidence after multi-hop attention routing.

\subsection{Cross-Frame Connectivity}
\label{appx:pre:cfc}
VTG requires comparing states before and after a boundary, which relies on attention paths that connect evidence across time \cite{abnar2020quantifying,chefer2021transformer}.
We quantify the cross-frame routing strength at layer $\ell$ by the total transition mass from frame $t$ to $t{+}1$:
\begin{equation}
\label{eq:cfc}
\Gamma^{(\ell)}(t)
= \sum_{u_i\in\mathcal{V}_t} \sum_{u_j\in\mathcal{V}_{t+1}} \mathbb{P}^{(\ell)}(u_i\!\to\!u_j),
\end{equation}
where $\mathcal{V}_t$ is the set of tokens in frame $t$.
Low $\Gamma^{(\ell)}(t)$ indicates weak temporal routing, meaning evidence has difficulty propagating across this boundary.

\section{Maximal Marginal Relevance (MMR) Algorithm}
\label{appx:mmr_algo}

In this section, we describe the algorithmic procedure of our Maximal Marginal Relevance (MMR) selection, which balances query relevance with token diversity for object evidence. 
The core idea is to suppress near-duplicate selections and only allow visually similar candidates when they introduce sufficiently different semantic information.

\begin{algorithm}[H]
\caption{MMR-based object token selection}
\label{alg:mmr_object_adjacent_simple}
\footnotesize
\KwIn{Normalized patch tokens $\{\hat{\mathbf{V}}_{patch}^{(t)}\in\mathbb{R}^{P\times D}\}_{t=1}^{T}$;
object budgets $\{k_{\text{obj}}^{(t)}\}_{t=1}^{T}$;
trade-off $\lambda_{mmr}\in[0,1]$;
temporal weight $\eta\in[0,1]$.}
\KwOut{Selected indices $\{\mathcal{S}^{(t)}\}_{t=1}^{T}$ with $|\mathcal{S}^{(t)}|=k_{\text{obj}}^{(t)}$.}

Compute patch-query similarities $s_{evi}$ by Eq.~(\ref{eq:obj_evi})\;
Initialize $\mathcal{S}^{(0)}\leftarrow\emptyset$\;

\For{$t=1$ \KwTo $T$}{
    Initialize candidates $\mathcal{C}\leftarrow\{1,\dots,P\}$ and $\mathcal{S}^{(t)}\leftarrow\emptyset$\;
    Initialize max-redundancy vector $m[p]\leftarrow 0,\ \forall p \in C$\;

    \For{$i=1$ \KwTo $k_{\text{obj}}^{(t)}$}{
        Select $p^\star \leftarrow \arg\max\limits_{p\in\mathcal{C}}
        \left(\lambda_{mmr}\, s_{evi}^{(t,p)}-(1-\lambda_{mmr})\, m[p]\right)$\;
        Update $\mathcal{S}^{(t)}\leftarrow \mathcal{S}^{(t)}\cup\{p^\star\}$ and $\mathcal{C}\leftarrow \mathcal{C}\setminus\{p^\star\}$\;

        Compute similarities between the newly selected token $p^\star$ and all candidates by a matrix--vector product:
        $\mathbf{s}=\hat{\mathbf{V}}_{patch}^{(t)}\,\hat{\mathbf{V}}_{t,p^\star}\in\mathbb{R}^{P}$\;
        Update the running max-sim redundancy:
        $m[p]\leftarrow \max\big(m[p],\, s[p]\big),\ \forall p\in\mathcal{C}$\;
    }
}
\Return{$\{\mathcal{S}^{(t)}\}_{t=1}^{T}$}\;
\end{algorithm}

A naive MMR implementation recomputes the redundancy term by comparing each candidate with all previously selected tokens, yielding $\mathcal{O}(T^2 \cdot P)$ cost, where T is the frame amount and P is the number of patches within each frame.
We reduce it to $\mathcal{O}(T \cdot P)$ by maintaining a running max-sim vector $m[p]$ for each candidate patch $p$.
After selecting a new token $p^\star$, we update $m$ incrementally using a single matrix-vector product and set $m\leftarrow \max(m,\mathbf{s})$.

\section{Additional Implementation Details}
\label{appx:implementation}
\subsection{Benchmarks}

\textbf{Charades-STA} \cite{sigurdsson2016hollywood} contains 3,720 testing queries over 1,334 videos of indoor activities with around 30 seconds per video.
The average sequence length after tokenization is 3,909 for Qwen3-VL and 5,322 for Qwen2.5-VL.

\noindent
\textbf{ActivityNet-Grounding} \cite{caba2015activitynet} contains 17,031 queries over 4,885 untrimmed videos, each of which has around 2 minutes describing both indoor and outdoor activities.
The average sequence length after tokenization is 10,460 for Qwen3-VL and 11,394 for Qwen2.5-VL.

\noindent
\textbf{Video-MME} \cite{fu2025video} The validation split of Video-MME contains 900 videos (its long-video subset averages about 40 minutes) with 2,700 multiple-choice QA pairs. Following prior settings \cite{shen2025fastvid}, we use the w/o subtitles pipeline.

\noindent
\textbf{LongVideoBench} \cite{wu2024longvideobench} consists of 752 long videos (up to 1 hour) and 1,337 questions across 17 fine-grained categories.

\subsection{Unified Protocol}
Evaluation pipelines for video-VLM pruning vary substantially (prompts, frame sampling, etc.), which can make comparisons misleading \cite{dart}.
We introduce \textbf{Open} \textbf{V}isual-\textbf{P}runing \textbf{S}uite \textbf{(OpenVPS)}, a unified evaluation protocol to enable fair comparisons for visual pruning.
To reduce evaluation cost, we disable the model's thinking-style generation by inserting the \texttt{</think>} token immediately after the prompt, following the settings in Time-R1 \cite{wang2025time}.
The prompts we used are followed by the Qwen3-VL \cite{yang2025qwen3}, as recorded in Appx. \ref{appx:sec:prompt_vtg} and \ref{appx:sec:prompt_vqa}.

For preprocessing, we resize videos by setting the shorter side to 480 pixels for Charades-STA~\cite{sigurdsson2016hollywood} and 256 pixels for ActivityNet~\cite{caba2015activitynet}, since ActivityNet clips are substantially longer.
This follows the standard practice in OpenTAD~\cite{liu2025opentad}. 
We use the default video loading pipeline of each base model. 
For Qwen-VL models, we sample raw frames at 2 FPS and apply $2{\times}2$ spatiotemporal merging, resulting in an effective visual input rate of 1 FPS. 
Qwen~\cite{bai2025qwen2} additionally employs dynamic resolution when the visual token length exceeds a preset threshold, which means the input resolution is automatically reduced.
We keep this threshold at 16,384 to match the vanilla setting. 
This setting with dense frame sampling aligns well with temporal grounding tasks.
For LLaVA-OneVision, we uniformly sample 32 frames for VideoQA evaluation, following prior work~\cite{shen2025fastvid}.

For VTG, following~\cite{ren2024timechat,zeng2024timesuite}, we report mean Intersection over Union (mIoU) and $R1$ at tIoU thresholds $m\!\in\!\{0.3,0.5,0.7\}$, i.e., the percentage of samples whose predicted segment attains IoU larger than $m$.
Auxiliary, we report accuracy for VideoQA in this appendix.
For efficiency, we estimate TFLOPs from the final sequence length and model sizes as detailed in Appx. \ref{appx:sec:tflops}.

All experiments are conducted on three NVIDIA L40 GPUs (48\,GB each).
Evaluation takes approximately 2 GPU-hours (single L40) on Charades-STA and 18 GPU-hours on ActivityNet.
Enabling thinking-style generation substantially increases the runtime while yielding only marginal gains in performance \cite{yang2025qwen3}.

\subsection{TFLOPs}
\label{appx:sec:tflops}
Following prior work \cite{fastv,xing2024pyramiddrop}, we estimate the theoretical Floating-point Operations Per Second (FLOPs) of the Video-VLMs.
Qwen \cite{bai2025qwen2,yang2025qwen3} uses Grouped-Query Attention (GQA) \cite{ainslie2023gqa} and a SwiGLU-based \cite{shazeer2020glu} three-layer feed-forward network.
Accordingly, the per-layer FLOPs of the LLM can be expressed as:
\begin{equation}
2nD(h_{kv}d) + 2nD^2 + 2n^2D + 3nDD',
\end{equation}
where $n$ denotes the number of video tokens, $D$ is the hidden size, $D'$ is the intermediate FFN width, $h_{kv}$ is the number of key/value heads, and $d$ is the head dimension.

\subsection{Instruction Prompts for Video Temporal Grounding}
\label{appx:sec:prompt_vtg}
\begin{tcolorbox}[
  colback=gray!3,
  colframe=black!25,
  arc=2pt,
  boxrule=0.4pt,
  left=6pt,right=6pt,top=6pt,bottom=6pt
]
\small
\ttfamily
Given a textual query: \textit{\{query\}}.\\
When does the described content occur in the video?\\
Please return the final timestamps in seconds directly in one sentence.
\end{tcolorbox}

\subsection{Instruction Prompts for Video Question Answering}
\label{appx:sec:prompt_vqa}
\begin{tcolorbox}[
  colback=gray!3,
  colframe=black!25,
  arc=2pt,
  boxrule=0.4pt,
  left=6pt,right=6pt,top=6pt,bottom=6pt
]
\small
\ttfamily
Select the best answer to the following multiple-choice question based on the video. 
Respond with only the letter of the correct option.\\
Question: \textit{\{query\}}\\  
Possible answer choices: \textit{\{choices\}}\\
The best answer is:
\end{tcolorbox}

\section{Additional Experiment}
\label{appx:exp}

\subsection{More Retention Ratio of Qwen2.5-VL-based SemVID on Video Temporal Grounding}

\begin{table*}[!hb]
\centering
\small
\setlength{\tabcolsep}{4pt}
\begin{adjustbox}{width=1\textwidth,center}
\begin{tabular}{l| cccc c |cc| cccc c |cc}
\toprule
\multicolumn{1}{l|}{} &
\multicolumn{7}{c|}{ActivityNet-Grounding} &
\multicolumn{7}{c}{Charades-STA} \\
\cmidrule(lr){2-8}\cmidrule(lr){9-15}
Method &
R1@0.3 & R1@0.5 & R1@0.7 & mIoU & (\%) & ER& CS &
R1@0.3 & R1@0.5 & R1@0.7 & mIoU & (\%) & ER& CS  \\
\midrule
\rowcolor{lightgray}
\multicolumn{15}{l}{\textbf{Qwen2.5-VL-7B, Retain 25\% Tokens}} \\
\midrule
Qwen2.5-VL-7B    &    
29.22 & 15.77 & 7.49 & 22.25 & 100 & $1e^{4}$ & 71.8 &
77.39 & 59.33 & 33.82 & 56.85 & 100 & $1e^{4}$ & 90.9 \\

FastV \cite{fastv} &
\multicolumn{5}{c|}{Out-of-Memory (OOM)} & -& - 
&\multicolumn{5}{c|}{Out-of-Memory (OOM)} & - & -\\
VScan \cite{zhang2025vscan} &
\multicolumn{5}{c|}{Out-of-Memory (OOM)} & - & - 
&\multicolumn{5}{c|}{Out-of-Memory (OOM)} & - & - \\

DART \cite{dart}          & 
17.45 & 9.11 & 4.25 & 13.56 & 60.9 & 3.4 & 20.5 &
44.41 & 30.32 & 15.46 & 30.11 & 52.9  & 3.6 & 22.0 \\

ToME \cite{bolya2022tome} & 
23.45 & 12.16 & 5.35 & 18.34 & 82.4 & 5.2 & 28.2 &
57.02 & 38.74 & 18.39 & 38.44 & 67.6 & 4.9 & 29.1 \\

TokenSculpt \cite{qatokensculpt} & 
23.42 & 12.10 & 5.34 & 18.41 & 82.7 & 5.5& 26.6 &
58.06 & 38.79 & 19.01 & 38.78 &68.2 & 5.1 & 28.1 \\

FastVID \cite{shen2025fastvid}     &
22.05 & 11.40 & 5.12 & 17.29 & 77.7 & 5.0 & 30.4 &
64.26 & 43.79 & 20.83 & 41.99 & 73.8 & 5.3 & 33.9  \\

VisionZip \cite{yang2025visionzip} & 
23.92 & 12.21 & 5.31 & 18.64 & 83.8 & 5.9 & 29.3 &
63.25 & 42.12 & 20.10 & 41.49 & 73.0 & 5.7 & 31.3 \\

\textbf{SemVID (ours)} & 
\textbf{25.06} & \textbf{13.07} & \textbf{5.85} & \textbf{19.50} & \textbf{87.7} & \textbf{6.1}& \textbf{33.1} &
\textbf{65.73} & \textbf{45.19} & \textbf{22.00} & \textbf{43.66} & \textbf{76.8}& \textbf{6.0} & \textbf{34.8} \\

\midrule
\rowcolor{lightgray}
\multicolumn{15}{l}{\textbf{Qwen2.5-VL-7B, Retain 33\% Tokens}} \\
\midrule
Qwen2.5-VL-7B   &    
29.22 & 15.77 & 7.49 & 22.25 & 100 & $1e^{4}$ & 71.8 &
77.39 & 59.33 & 33.82 & 56.85 & 100 & $1e^{4}$ & 90.9 \\

FastV \cite{fastv} &
\multicolumn{5}{c|}{Out-of-Memory (OOM)} & -& - 
&\multicolumn{5}{c|}{Out-of-Memory (OOM)} & - & -\\
VScan \cite{zhang2025vscan} &
\multicolumn{5}{c|}{Out-of-Memory (OOM)} & - & - 
&\multicolumn{5}{c|}{Out-of-Memory (OOM)} & - & - \\

DART \cite{dart}          & 
18.62 & 9.64 & 4.39 & 10.88 & 48.9 & 1.9 & 13.6 &
49.81 & 33.31 & 17.45 & 33.53 & 59.0& 3.9 & 22.8 \\

ToME \cite{bolya2022tome} & 
24.13 & 12.56 & 5.35 & 18.84 & 84.7 & 6.1 & 34.3 &
62.15 & 42.58 & 20.13 & 41.39 & 72.8& 5.5 & 35.2 \\

TokenSculpt \cite{qatokensculpt} & 
24.95 & 12.90 & 5.69 & 19.40 & 87.2 & 6.7& 36.1 &
65.43 & 45.32 & 22.77 & 43.53 & 76.6  & 6.2 & 35.7 \\

FastVID \cite{shen2025fastvid}     &
24.82 & 12.66 & 5.51 & 19.23 & 86.4 & 6.2 & 38.2 &
69.41 & 49.46 & 24.57 & 45.98 & 80.9  & 6.4 & 39.6  \\

VisionZip \cite{yang2025visionzip} & 
25.19 & 12.92 & 5.75 & 19.48 & 87.5 & 6.8 & 36.0 &
67.58 & 47.69 & 23.47 & 44.87 & 78.9 & 6.4 & 36.7 \\

\textbf{SemVID (ours)} & 
\textbf{25.22} & \textbf{13.06} & \textbf{5.79} & \textbf{19.63} & \textbf{88.3} & \textbf{6.9}& \textbf{39.2} &
\textbf{69.39} & \textbf{49.23} & \textbf{24.68} & \textbf{46.12} & \textbf{81.1} & \textbf{6.7} & \textbf{40.3} \\

\bottomrule
\end{tabular}
\end{adjustbox}
\caption{VTG results with Qwen2.5-VL under 25\% and 33\% retention. FastV and VScan materialize full self-attention matrix and lead to OOM in long videos.}
\label{tab:appx_qw2_vtg}
\end{table*}

\cref{tab:appx_qw2_vtg} shows that SemVID remains effective under different pruning rates. 
On both Charades-STA and ActivityNet with Qwen2.5-VL, SemVID consistently delivers strong VTG performance at 25\% and 33\% token retention, achieving the best mIoU scores and performance maintenance under the same budget. 
These results indicate that our semantic evidence allocation is not tuned to a single compression point but provides a stable accuracy-efficiency trade-off across a wide range of token budgets.

\begin{table*}[t]
\centering
\small
\setlength{\tabcolsep}{4pt}
\begin{adjustbox}{width=0.97\textwidth,center}
\begin{tabular}{l| cc | cc| cc| cc}
\toprule
\multicolumn{1}{l|}{} &
\multicolumn{2}{c|}{Motion $\uparrow$, Transition $\downarrow$}&
\multicolumn{2}{c|}{Motion $\uparrow$, Transition $\uparrow$}&
\multicolumn{2}{c|}{Motion $\downarrow$, Transition $\downarrow$}&
\multicolumn{2}{c}{Motion $\downarrow$, Transition $\uparrow$} \\
\cmidrule(lr){2-9}
Method  & mIoU & (\%) & mIoU & (\%)& mIoU & (\%)& mIoU & (\%)\\
\midrule
Qwen3-VL-4B                        &
 46.9 & 100 &    37.7 & 100 &    42.7 & 100 &    30.2 & 100 \\
FastVID \cite{shen2025fastvid}     &
 39.6 & 84.4 &   33.8 & 89.7 &   34.7 & 81.3 &   24.7 & 81.8 \\
\textbf{SemVID (ours)} &
\textbf{46.5}&\textbf{99.3}&   \textbf{35.1}&\textbf{93.1}&   \textbf{41.3}&\textbf{96.7}&   \textbf{26.9}&\textbf{89.1}\\

\bottomrule
\end{tabular}
\end{adjustbox}
\caption{Sensitivity analysis of motion amplitude and background transition presence under 25\% retention ratio.}
\label{tab:appx:motion_amplitude}
\end{table*}

\subsection{Sensitivity Analysis of Motion Amplitude}
\label{appx:exp:motion_amplitude}

In this section, we examine whether our motion tokens remain effective when the target motion is subtle and the video contains strong background or camera transition artifacts.
To identify strong background or camera-induced transition artifacts, we run PySceneDetect, which computes an inter-frame content-difference score based on frame appearance changes in HSV space and flags a scene boundary whenever the score exceeds a threshold. 
We set the threshold to a relatively low value of 15 so that camera motion is more likely to be detected as a scene change.
For motion amplitude, we use GPT-4o-mini \cite{openai_gpt4o} to classify each query into small amplitude, large amplitude, or not sure, where the not sure category is used to alleviate potential bias or noise. 
We then randomly sample instances from the Charades dataset and stratify them into four bins formed by the cross-product of motion amplitude and background transition presence. 
Sampling continues until each bin contains 100 instances, yielding 400 instances in total. 
This balanced design mitigates distributional skew and enables a direct comparison of motion-token behavior in the most challenging regime, namely small motions under strong background transitions.

As shown in \cref{tab:appx:motion_amplitude}, VTG accuracy is affected more by background transition artifacts than by action amplitude. 
For the original setting, introducing strong transitions consistently causes a large mIoU drop (46.9$\rightarrow$37.7 and 42.7$\rightarrow$30.2), indicating that background or camera changes inject substantial noise into temporal grounding.

FastVID~\cite{shen2025fastvid} mitigates part of this issue by preserving tokens mainly from sparse anchor frames, which reduces exposure to transition-heavy frames and thus narrows the discrepancy between transition-present and transition-absent cases. 
However, VTG fundamentally relies on temporally adjacent state changes around boundaries.
Aggressively compressing boundary-adjacent moments weakens transition cues and leads to a clear mIoU degradation across all settings.

In contrast, SemVID remains consistently strong under all motion and transition regimes. 
While our coarse allocation may assign quotas to irrelevant transitions, the subsequent query-filtered motion-token selection suppresses background-dominated changes and prioritizes semantically meaningful motion evidence. 
This yields uniformly high retention, reaching 99.3\% in the easiest setting and maintaining 89.1\% even in the most challenging case (Motion $\downarrow$, Transition $\uparrow$), demonstrating improved robustness to both subtle actions and strong background transition artifacts.

\subsection{Performance of Qwen2.5-VL-based SemVID on Video Question-Answering}
\begin{table*}[t]
\centering
\small
\setlength{\tabcolsep}{4pt}
\begin{adjustbox}{width=0.85\textwidth,center}
\begin{tabular}{l l| cc| cc| cccc c| cc}
\toprule
\multicolumn{2}{l|}{} &
\multicolumn{2}{c|}{Tokens} &
\multicolumn{2}{c|}{TFLOPs} &
\multicolumn{7}{c}{VideoMME} \\
\cmidrule(lr){3-4}\cmidrule(lr){5-6}\cmidrule(lr){7-13}
Method & Size & \# & (\%) & Value & (\%) &
Short & Medium & Long & Overall & (\%) & ER & CS\\
\midrule
Qwen2.5-VL-7B    & 7B & 13447 & 100 & 124.0 & 100 &
76.7 & 68.2 & 56.9 & 67.3 & 100 &
$1e^4$ & 66.3\\

PruneVID \cite{huang2025prunevid}  & 7B & 3295 & 24.5 & 23.7 & 19.1 &
67.0 & 59.4 & 51.6 & 59.3 & 88.1 &
- & -\\

FastVID \cite{shen2025fastvid}     & 7B & 3240 & 24.1 & 23.2 & 18.7 &
\textbf{74.3} & \underline{61.3} & \textbf{52.8} & \underline{62.8} & \underline{93.3} &
5.2 & \textbf{33.0}\\

\textbf{SemVID}                    & 7B & 3222 & 24.0 & 23.1 & 18.7 &
\underline{73.8} & \textbf{62.8} & \underline{52.4} & \textbf{63.1} & \textbf{93.7} &
\textbf{5.4} & 30.7\\

\bottomrule
\end{tabular}
\end{adjustbox}
\caption{VQA results on VideoMME with Qwen2.5-VL-7B under 24\% retention.}
\label{tab:main_videomme_qw2}
\end{table*}

Although SemVID is primarily developed for VTG, it transfers effectively to general VideoQA. 
As shown in \cref{tab:main_videomme_qw2}, under a comparable token budget, SemVID reaches 63.1\% overall accuracy on VideoMME with Qwen2.5-VL-7B, outperforming representative pruning baseline PruneVID and yielding a modest but consistent gain over FastVID. 
Notably, SemVID also achieves higher Evidence Retention (ER) than FastVID, suggesting that preserving structured evidence (diverse object evidence, query-filtered motion cues, and context anchors) benefits not only boundary localization but also general perception tasks. Meanwhile, the relatively smaller role of Connectivity Strength (CS) for VideoQA is expected since many questions can be resolved from one or a few informative frames rather than requiring a temporally continuous evidence chain.

Following the evaluation protocol of FastVID~\cite{shen2025fastvid}, which reports Qwen2.5-VL results only on VideoMME, we restrict Qwen2.5-VL comparisons to VideoMME to ensure a consistent and fair benchmark.

\subsection{Performance of LLaVA-OneVision-based SemVID on Video Question-Answering}
\begin{table*}[t]
\centering
\small
\setlength{\tabcolsep}{4pt}
\begin{adjustbox}{width=0.79\textwidth,center}
\begin{tabular}{l l|c| cc| cccc c}
\toprule
\multicolumn{2}{l|}{} &
\multicolumn{1}{c|}{Tokens} &
\multicolumn{2}{c|}{LongVideoBench} &
\multicolumn{5}{c}{VideoMME} \\
\cmidrule(lr){3-3}\cmidrule(lr){4-5}\cmidrule(lr){6-10}
Method & Size & (\%) & Value & (\%) &
Short & Medium & Long & Overall & (\%) \\
\midrule
LLaVA-OneVision-7B    & 7B  & 100 & 56.6 & 100 & 70.1 & 56.6 & 48.8 & 58.5 & 100 \\

FastV \cite{fastv}  & 7B  & 25.0 &
\textbf{56.8} & \textbf{100.4} &
66.0 & 54.6 & 47.2 & 55.9 & 95.6 \\

VisionZip \cite{yang2025visionzip}     & 7B & 25.0 &
56.0 & 98.9 &
68.9 & 57.4 & 47.6 & 58.0 & 99.1 \\

PruneVid \cite{huang2025prunevid}     & 7B & 25.0 &
55.1 & 97.3 &
68.8 & 54.4 & 47.7 & 57.0 & 97.4 \\

FrameFusion \cite{fu2025framefusion}     & 7B & 25.0 &
54.8 & 96.8 &
68.2 & 55.7 & 48.6 & 57.5 & 98.3 \\

FastVID \cite{shen2025fastvid}     & 7B & 25.0 &
56.3 & 99.5 &
\textbf{69.9} & 56.6 & 47.7 & 58.0 & 99.1 \\

\textbf{SemVID}                    & 7B & 25.0 &
56.4 & 99.7 &
68.7 & \textbf{57.0} & \textbf{48.7} & \textbf{58.1} & \textbf{99.4} \\
\midrule
LLaVA-OneVision-7B  & 7B  & 100 & 56.6 & 100 & 70.1 & 56.6 & 48.8 & 58.5 & 100 \\

FastV \cite{fastv}  & 7B  & 15.0 &
51.5 & 91.0 &
58.4 & 51.7 & 45.4 & 51.9 & 88.7 \\

VisionZip \cite{yang2025visionzip}     & 7B & 15.0 &
54.2 & 95.8 &
63.8 & 54.4 & 48.3 & 55.5 & 94.9 \\

PruneVid \cite{huang2025prunevid}     & 7B & 15.0 &
55.6 & 98.2 &
67.9 & 54.3 & 48.1 & 56.8 & 97.1 \\

FrameFusion \cite{fu2025framefusion}     & 7B & 15.0 &
53.0 & 93.6 &
65.8 & 54.1 & 46.7 & 55.5 & 94.9 \\

FastVID \cite{shen2025fastvid}     & 7B & 15.0 &
56.2 & 99.3 &
\textbf{69.3} & 56.2 & 47.4 & 57.7 & 98.6 \\

\textbf{SemVID}                    & 7B & 15.0 &
\textbf{56.4} & \textbf{99.7} &
68.3 & \textbf{56.7} & \textbf{48.7} & \textbf{57.9} & \textbf{99.0} \\

\bottomrule
\end{tabular}
\end{adjustbox}
\caption{VQA results on LongVideoBench and VideoMME with LLaVA-OneVision-7B under 15\% and 25\% retention.}
\label{tab:main_vqa_llavaov}
\end{table*}

LLaVA-OneVision is a widely used multimodal model with strong image-level reasoning and a pragmatic video interface, making it a common choice in recent visual token pruning and efficiency-oriented studies. 
However, LLaVA-OneVision is not intended for fine-grained temporal localization. 
In its standard inference pipeline, videos are represented by a fixed, sparse set of uniformly sampled 32 frames, and the model does not provide an explicit mechanism to perceive or output timestamps. 
This sampling strategy is sufficient for general VideoQA tasks whose answers can be supported by object-centric cues from key frames, but it inevitably discards dense motion patterns and long-range temporal dependencies, and therefore no previous study applies LLaVA-OneVision on VTG.

To align the evaluation with the backbone's strengths and ensure a fair comparison, we focus on VideoQA for LLaVA-OneVision. 
Moreover, since motion cues are underrepresented under sparse frame sampling (32 frames only for a video of hours length), we set $\alpha=0.9$ in Eq.~(\ref{eq:ba_w}) and $\lambda_{\text{mmr}}=0.5$ in Eq.~(\ref{eq:sti_mmr}) to place greater emphasis on object-centric evidence while improving the diversity of selected semantic cues. 
This also demonstrates the robustness of our method across different downstream tasks, as it can be readily adapted by simply adjusting the priority of object and motion tokens.

According to \cref{tab:main_vqa_llavaov}, the advantage of our SemVID is most evident on long videos, largely owing to our designs for the diversity (MMR-based object selection) and spatially-temporally structure preserving (semantic budget allocation). 
In long-form videos, many pruning strategies implicitly concentrate tokens on a few highly salient or query-matched regions, which leads to evidence collapse.
By contrast, SemVID explicitly selects diverse semantic evidence and spreads tokens across different objects in different scenes, yielding a compact yet richer evidence pool that better supports long-range aggregation.

\subsection{Visualization to Our Semantic Budget Allocation}

\begin{figure}[htbp]
  \centering
  \includegraphics[width=0.95\textwidth]{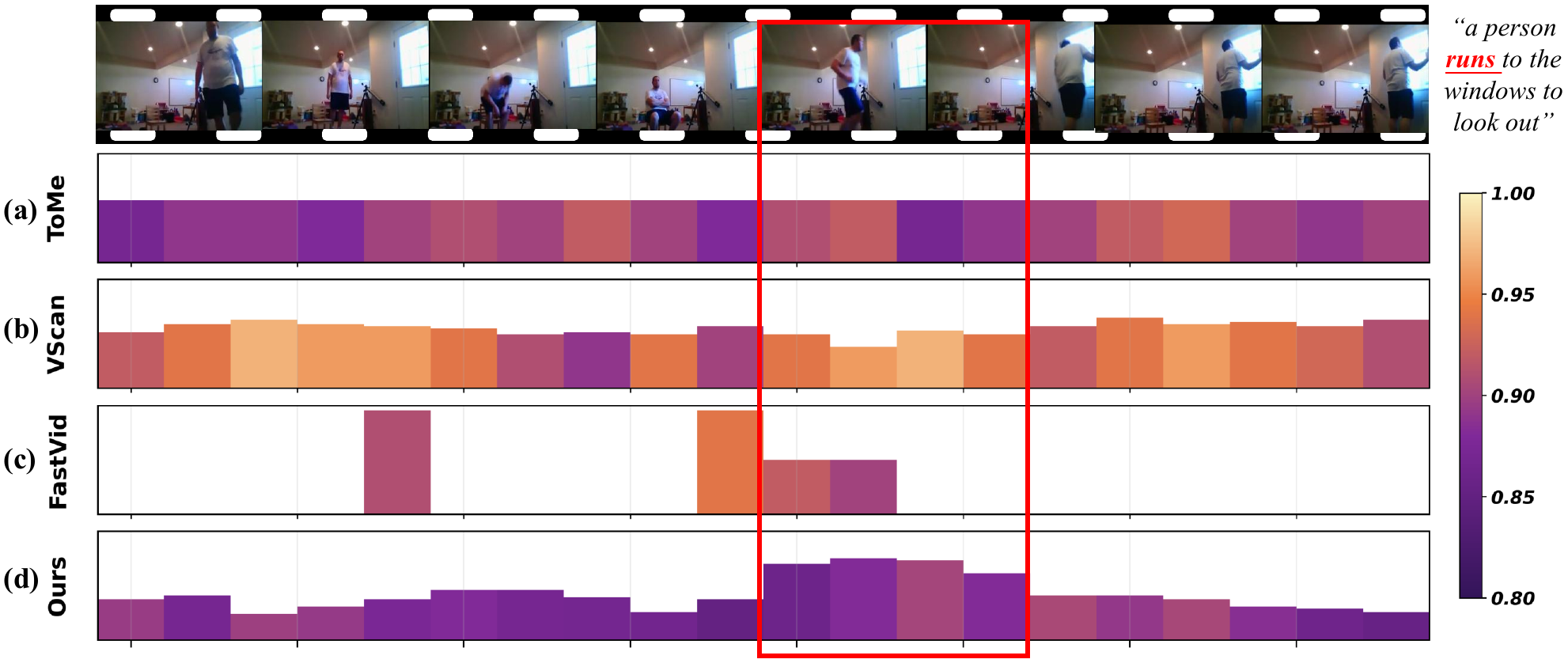}
\caption{Per-frame token allocation in VTG.
Bar height indicates the \textbf{number of retained tokens per frame}; color denotes the average similarity between tokens retained in the current frame and all retained tokens (\textbf{darker = more diverse and less redundant}); the red box marks the ground-truth moment. We compare ToMe~\cite{bolya2022tome}, VScan~\cite{zhang2025vscan}, FastVID~\cite{shen2025fastvid}, and SemVID (ours).}
  \label{fig:vis_budget_allocation}
\end{figure}

\cref{fig:vis_budget_allocation} reveals that where and how the retained tokens are distributed over time is essential.
\textbf{ToMe} \cite{bolya2022tome} uniformly pruning spreads tokens over \textbf{redundant} frames, under-emphasizing boundary-critical moments. \textbf{VScan} \cite{zhang2025vscan} relies on \textbf{relevance}-driven selection, which repeatedly picks the same query-related regions across frames, collapsing diversity and missing transitions. 
\textbf{FastVID} \cite{shen2025fastvid} employs anchor-based aggregation on \textbf{salient} patches, sparsifying the timeline and creating token-empty gaps that break cross-frame continuity. 
In contrast, our \textbf{SemVID} explicitly assigns budget according to the \textbf{semantic} role of evidence, resulting in a more structured timeline. 
It increases token density around the ground-truth interval to preserve boundary-critical cues while still reserving a non-trivial budget for intermediate frames.

Crucially, the intermediate allocation is not wasted redundancy.
It retains motion and context semantics that serve as bridges between temporally separated object evidence, preventing evidence from becoming fragmented and improving cross-frame connectivity for multi-hop reasoning.
This budget allocation yields a compact yet diverse evidence set (darker colors) without creating token-empty gaps, which better supports both evidence retention near boundaries and coherent propagation across frames.

\subsection{Visualization of the Evidence Retention}

\begin{figure}[htbp]
  \centering
  \includegraphics[width=0.87\textwidth]{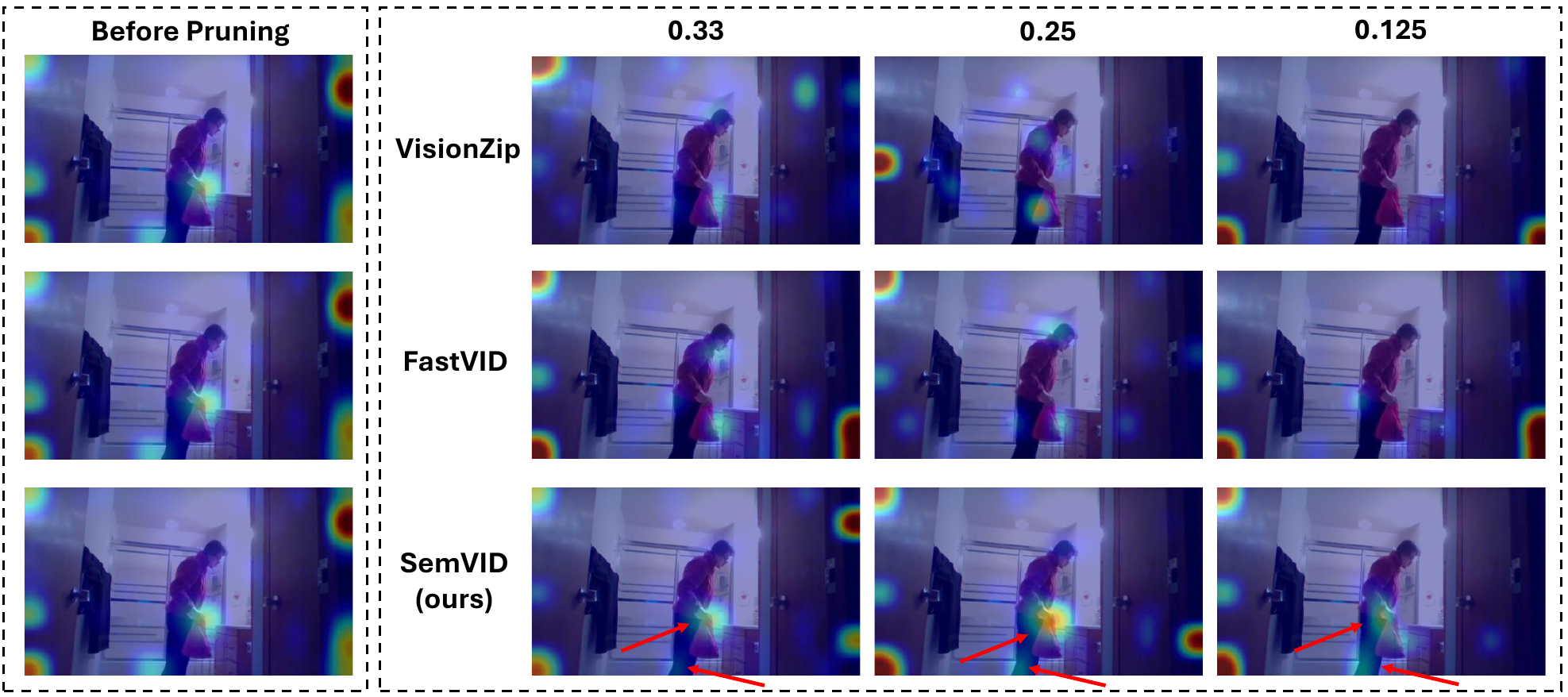}
\caption{Comparisons of the attention landing distribution $\boldsymbol{\pi}^{(1)}$ (Eq.~\ref{eq:pi_layer}) over input patches for the query \textit{"the person takes a bag from the bottom cabinet"} on Qwen3-VL-4B, shown under different token retention ratios.}
  \label{fig:vis_evidence_retention}
\end{figure}

The attention landing distribution $\boldsymbol{\pi}^{(1)}$ is derived by injecting query evidence through the last-layer cross-attention and propagating it backward via cross-layer self-attention over the vision token graph.
The resulting distribution at the first layer, referred to as the landing distribution, highlights patches that are most critical to the query.
Implementation details are provided in Appx.~\ref{appx:pre:cfc}.

\cref{fig:vis_evidence_retention} visualizes the effect of our evidence retention objective.
SemVID first selects object-centric tokens to preserve query-critical evidence and further introduces motion tokens to capture foreground semantic transitions, which jointly encourage the attention mass to concentrate on the true evidence regions.
Notably, SemVID assigns substantially higher attention to boundary-defining cues such as the red bag and the hands while preserving a large fraction of the pre-pruning attention mass (highlighted by red arrows), whereas other pruning strategies tend to under-attend these regions.

\subsection{Detailed Analysis on Efficiency}

\begin{table}[H]
\centering
\setlength{\tabcolsep}{2.3pt}
\resizebox{0.75\columnwidth}{!}{
\begin{tabular}{lcccc}
\toprule
\multirow{2}{*}{Module} &
\multicolumn{2}{c}{Short videos} &
\multicolumn{2}{c}{Long videos} \\
\cmidrule(lr){2-3}\cmidrule(lr){4-5}
& VRAM (GB) & Lat. (ms) & VRAM (GB) & Lat. (ms) \\
\midrule
Obj. token: Eq.~(5) only               & 2.8 & 13.0 & 7.6 & 28.3 \\
Obj. token: Eq.~(5) + MMR              & 2.8 & 16.2 & 7.6 & 39.9 \\
Motion token                           & 1.5 & 1.4 & 4.4 & 3.4 \\
Context token                         & 0.3 & 0.7 & 1.2 & 0.9 \\
\midrule
Attn.-based pruning      & 28.8 & 847.8 & OOM & OOM\\
Similarity-based pruning & 3.4 & 10.8 & 17.1 & 36.8 \\
\bottomrule
\end{tabular}}
\caption{Independent profiling of pruning overhead.}
\label{tab:appx_efficiency}
\end{table}

\cref{tab:appx_efficiency} provides component-wise profiling of VRAM (pruning only w/o raw tokens) and latency on Short/Long splits obtained by dividing ActivityNet into equal-size groups by duration.
It shows that our accelerated MMR (Appx. \ref{appx:mmr_algo}) introduces only $\sim$1\% VRAM overhead.

Although SemVID introduces overhead from patch-query similarity and MMR, its memory scaling is substantially lighter than existing alternatives.
Attention-based FastV derives token importance from full attention over visual tokens, and similarity-based FastVID relies on token-token similarity to remove redundancy.
Both can introduce large intermediate maps with quadratic scaling up to $(TP)^2$.
In contrast, \cref{eq:obj_evi} only computes query-patch relevance and reduces it to a $TP$-length vector, making it more practical for long-video inference.

\end{document}